\documentclass[letterpaper]{article}

\usepackage[round]{natbib}  
\usepackage{alifeconf}
\usepackage{url,hyperref}
\usepackage{microtype}
\usepackage{subcaption}

\usepackage{amsmath}
\usepackage{amssymb}
\usepackage{mathtools}
\usepackage{amsthm}

\usepackage[capitalize,noabbrev]{cleveref}

\theoremstyle{plain}

\theoremstyle{definition}

\theoremstyle{remark}

\usepackage[disable,textsize=tiny]{todonotes}

\usepackage{amsmath,amsfonts,bm}

\def\eqref#1{equation~\ref{#1}}

\def\1{\bm{1}}

\def\vx{{\bm{x}}}

\DeclareMathAlphabet{\mathsfit}{\encodingdefault}{\sfdefault}{m}{sl}
\SetMathAlphabet{\mathsfit}{bold}{\encodingdefault}{\sfdefault}{bx}{n}

\def\sR{{\mathbb{R}}}

\usepackage{wrapfig}
\usepackage{hyperref}

\usepackage[utf8]{inputenc} 
\usepackage[T1]{fontenc}    
\usepackage{hyperref}       
\usepackage{url}            
\usepackage{booktabs}       
\usepackage{amsfonts}       
\usepackage{nicefrac}       
\usepackage{microtype}      
\usepackage{xcolor}         
\usepackage{xurl}
\usepackage{xspace} 
\usepackage{graphicx}
\usepackage{listings}
\usepackage{xcolor}
\usepackage{algorithm}
\usepackage{algorithmic}
\usepackage{comment}
\usepackage{lipsum}
\usepackage{tikz}
\usepackage[most]{tcolorbox}
\usepackage{amssymb}  
\usepackage{booktabs} 
\usepackage{caption} 
\usepackage{multirow} 
\usepackage{colortbl} 
\usepackage{longtable} 

\newcommand{\figremovespace}{\vskip -2pt}

\definecolor{populationcolor}{RGB}{0,128,0}     
\definecolor{pollutioncolor}{RGB}{220,20,60}    
\definecolor{resourcecolor}{RGB}{30,144,255}    
\lstdefinestyle{prompt}{
    language={},
    basicstyle=\small\ttfamily,
    numbers=none,
    frame=none,
    backgroundcolor={},
    escapeinside={(*}{*)},
    breakindent=0pt,
    showstringspaces=false,
    breaklines=true
}

\lstdefinestyle{goal}{
    language={},
    basicstyle=\small\ttfamily,
    numbers=none,
    frame=none,
    backgroundcolor={},
    breakindent=0pt,
    showstringspaces=false,
    breaklines=true
}

\lstdefinestyle{code}{
    language=Python
}

\definecolor{editcolor}{HTML}{AA3300}   
\definecolor{delcolor}{HTML}{9A9A9A}    

\newcommand{\method}{CEDAR\xspace}
\newcommand{\methodlong}{Complex-systems Exploration and Design via Agent-Orchestrated Refinement\xspace}

\newcommand\blfootnote[1]{%
  \begingroup
  \renewcommand\thefootnote{}\footnote{#1}%
  \addtocounter{footnote}{-1}%
  \endgroup
}

\title{CEDAR: Agent-Orchestrated Tree Search for Goal-Directed \\ Optimization of Complex Systems}

\author{
    Yingtao Tian$^{1}$ \\
    \mbox{}\\
    $^1$Sakana AI, Tokyo, Japan \\
    alantian@sakana.ai
}

\begin{document}

\maketitle


\begin{abstract}

Complex systems, core objects of study in artificial life, model diverse phenomena through nonlinear, feedback-driven interactions that produce emergent behavior, with applications from population dynamics and biology to economic policy and strategic decision-making.
Yet the difficulty of predicting how feedback structure gives rise to emergent behavior, a central open problem in artificial life, makes goal-directed design exceptionally challenging.
In established practice, system structures are written in specialized modeling languages such as DYNAMO or STELLA, compounding the challenge with labor-intensive workflows that limit adoption and hinder timely decision-making.

To address these challenges, we introduce \method, an autonomous method that uses Large Language Model (LLM) agents to discover complex systems satisfying user-specified behavioral goals.
Our key innovation is an LLM-driven Monte Carlo Tree Search (MCTS) deeply coupled with complex systems: at each iteration, an LLM Judge evaluates emergent behavior against specified goals and an LLM Editor proposes improved variants, with the Judge acting as a fitness function and the Editor as a variation operator, akin to a generate-and-evaluate loop in evolutionary computation.
We represent complex systems as a restricted, runnable subset of Python with domain-specific primitives, letting LLMs modify system dynamics directly.
\method formalizes this as an MCTS variant with an LLM-parameterized transition kernel and value function, enabling goal-directed discovery of complex system behaviors while preserving solution diversity, and its LLM-based interpretability reveals how structural changes drive emergent behavior.
\method reduces human effort while enabling capabilities difficult to achieve with existing approaches, facilitating broader adoption of complex systems across domains.

\end{abstract}


\blfootnote{\textcopyright 2026 Yingtao Tian. Published under a Creative Commons Attribution 4.0 International (CC BY 4.0) license.}

\section{Introduction}

Computational system modeling provides a framework for studying real-world phenomena as nonlinear, feedback-driven \emph{complex systems}, computational models whose global behavior emerges from feedback interactions among components~\citep{radzicki2008origin,richmond1985stella}.
Such systems are central objects of study in artificial life~\citep{bedau2003artificial,langton2019artificial,gershenson2023emergence} and systems thinking research~\citep{anderson1997systems}. This framework has been applied in various domains such as global dynamics~\citep{forrester1971world}, pandemic diffusion~\citep{zhu2025ekf}, social simulations~\citep{kolson1996politics}, and biological systems~\citep{hannon2014modeling}, critically supporting policy design and strategic decision-making~\citep{peterson2003barry,schunemann2024complex}. 
Yet predicting how feedback structure gives rise to emergent behavior, a central open problem in artificial life, remains notoriously difficult, making goal-directed design of such systems exceptionally challenging.

A further challenge is practical: established practice often involves specialized modeling languages such as DYNAMO~\citep{radzicki2008origin} and STELLA~\citep{richmond1985stella}, which require extensive manual effort to navigate unclear feedback relationships~\citep{guneralp2004principle} and labor-intensive workflows~\citep{sterman2000business}, limiting adoption across domains where complex systems modeling could provide valuable insights.
Large language models (LLMs) and LLM-powered agents now offer new capabilities for automated discovery of complex systems.
In particular, when combined with Monte Carlo Tree Search (MCTS)~\citep{kocsis2006bandit}, LLMs show strong capabilities in planning~\citep{zhao2023large,gao2024interpretable} and reasoning~\citep{song2025rekgmcts,chi2025thoughtsculpt}.
This opens promising directions for searching over spaces of complex systems, including artificial life discovery~\citep{kumar2025automating,nisioti2024text}, automated scientific discovery~\citep{yamada2025ai}, and evolving models to generate and falsify hypotheses about biological dynamics~\citep{srinivasan2025evolving}.

We propose \method{} (\methodlong), an autonomous method for discovering complex systems that meet specified goals by combining a unified system representation, the LLM Judge and Editor, and MCTS.
At each iteration, MCTS selects a system variant for expansion, an LLM Editor generates improved variants, and the variants are executed and evaluated by an LLM Judge.
Concretely, the LLM Editor acts as a variation operator proposing new system structures~\citep{lehman2023evolution}, while the LLM Judge serves as a fitness function evaluating each candidate, casting system discovery as an evolutionary search process guided by MCTS, where the search tree plays the role of a structured candidate population.
This is enabled by our unified representation using a restricted Python subset with domain-specific primitives and inline documentation.

In sum, \method{} enables goal-directed discovery of complex system behaviors with capabilities difficult to achieve with existing approaches, while preserving solution diversity that supports open-ended exploration, demonstrates applicability on classical systems drawn from social and biological modeling, and makes the emergent mechanisms of complex systems more transparent through LLM-based interpretability, lowering the barrier to constructing and studying complex systems in artificial life and beyond.

\begin{figure*}[t]
  \figremovespace
  \begin{center}
    \includegraphics[width=0.83\textwidth,trim=0 60pt 0 0,clip]{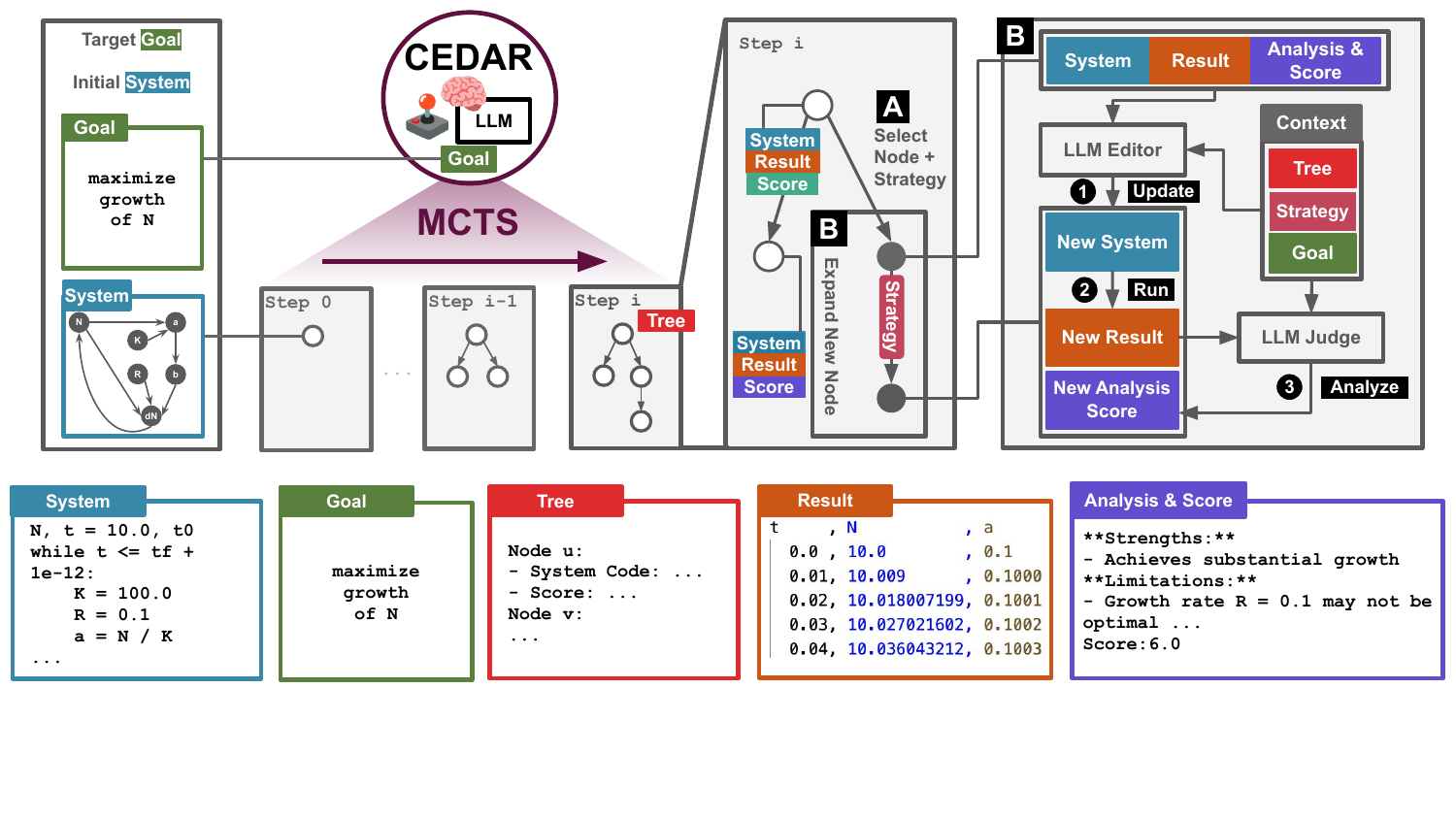}
    \caption{Overall architecture of \method. The autonomous approach uses Monte Carlo Tree Search (MCTS) to iteratively select system variants for expansion. At each iteration, an LLM Editor generates improved system variants and executes them, then an LLM Judge evaluates their performance and provides detailed analysis. The method integrates several components: a system in a restricted subset of Python code, a natural-language goal description, a search tree structure, system execution results, and comprehensive analysis feedback.}  
    \label{figure:mcts-diagram}
  \end{center}
  \figremovespace
\end{figure*}

{

\section{Related Work}
\label{section:related-works}

\paragraph{Complex Systems and How to Model Them}
\emph{World Dynamics}~\citep{forrester1971world} simulates population, resources, and industrial output over long horizons of the planet.
This introduces a computational framework for modeling a subject as a nonlinear, feedback-driven system (``complex system''),
a paradigm shift~\citep{forrester2011mit} from linear prediction-based modeling.
With the help of DYNAMO~\citep{radzicki2008origin} and STELLA~\citep{richmond1985stella} modeling languages (online examples of \href{https://github.com/bfix/dynamo/tree/master/rt}{DYNAMO} and \href{https://en.wikipedia.org/wiki/File:Cat_population,_STELLA_model.svg}{STELLA} show their old syntax and proprietary visual interfaces), complex systems have seen a wide range of applications, including social simulations~\citep{kolson1996politics}, war games, pandemic diffusion~\citep{zhu2025ekf}, economics, thermodynamics, and biological systems~\citep{ruth2012modeling,hannon2014modeling}.
This systems thinking~\citep{anderson1997systems,arnold2015definition} supports policy design and strategic decision-making~\citep{peterson2003barry,schunemann2024complex}.
Despite these advances, modeling still requires expert-defined variables and relationships~\citep{ford1998expert,berard2010gmb,zagonel2002model}: the relationship between structure and emergent behavior is often unclear~\citep{schoenberg2019understanding,guneralp2004principle,barlas1996formal}, and established languages require labor-intensive workflows~\citep{mit1998roadmaps,hines1996molecules,sterman2000business}.

\paragraph{Tree Search Algorithms with LLMs}
Monte Carlo Tree Search (MCTS)~\citep{kocsis2006bandit} has shown promising results when combined with powerful models.
MCTS with neural networks has led to superhuman performance in games~\citep{silver2016mastering,silver2017mastering,silver2017alphazero,schrittwieser2020mastering}, plus the discovery of novel algorithms~\citep{fawzi2022discovering,mankowitz2023faster} and neural network architectures~\citep{nasir2024llmatic}.
More recently, MCTS has been combined with large language models (LLMs), including LLM-powered value functions and self-reflection~\citep{zhou2024language}, LLM-powered node selection and expansion strategies~\citep{inoue2025wider},
as well as applications to specific tasks such as automatic design~\citep{zheng2025monte} and agent collaboration~\citep{gan2025master}.
Notably, MCTS with LLMs offers strong capabilities in planning, where LLMs serve as common-sense policy priors~\citep{zhao2023large} and enhance interpretability~\citep{gao2024interpretable}, and in reasoning, where LLMs guide traversals of knowledge graphs and refine reasoning through steps~\citep{song2025rekgmcts,chi2025thoughtsculpt}.

\paragraph{LLMs as Evolutionary Operators in Artificial Life}
LLMs have been proposed as variation operators in evolutionary search, replacing random mutation with language-guided program edits~\citep{lehman2023evolution,nisioti2024text}.
A related line evolves language prompts to steer the behavior of simulated cellular and biological systems~\citep{le2025giving,le2025zapgpt}.
This connects LLMs to a central ALife goal: open-ended discovery of emergent behaviors.
Our work extends this to complex systems, with the LLM Editor as variation operator and the LLM Judge as fitness function.

\paragraph{LLM-based Optimization for Structured Systems}
LLMs with MCTS show promising results for complex and challenging searches,
including automated scientific discovery~\citep{yamada2025ai},
artificial life discovery~\citep{kumar2025automating} and runnable code for reasoning~\citep{katz2024thought}.
Recent works also apply LLMs specifically to model system dynamics~\citep{liu2024llms,luo2025llm,liu2025leveraging} by treating models as predictors of behaviors that are \emph{unknown} to the optimizer. 
These studies consider only smaller systems (up to 4 variables and 12 steps), whereas our method scales to 60 variables and 4000 steps, making direct extensions of them to our scenario non-trivial.

\paragraph{Prior Works Close to Our Method}

Several prior works combine LLMs with search but target different goals.
I-MCTS~\citep{liang2025mcts} targets AutoML hyperparameters and LATS~\citep{zhou2024language} operates on discrete task graphs, and neither addresses iterative temporal dynamics optimization.
GIF-MCTS~\citep{dainese2024generating} and LLM-SRBench~\citep{shojaee2025llmsrbench} address short sequences, whereas our method optimizes multi-thousand-step temporal dynamics.
Prompt optimization methods~\citep{tong2025evoprompt,fernando2024promptbreeder,khattab2024dspy} are applicable in principle but require demonstrations unavailable in our online-exploration setting.
}

{

\section{Method}
\label{sec:method}

\subsection{Preliminaries}
We first formalize the complex systems under discussion in this work.
Complex systems model a subject as a nonlinear, feedback-driven mathematical system whose dynamics give rise to emergent behavior.
While such systems admit a range of formalisms, including agent-based models, cellular automata, and discrete-time dynamic graphs, one common formalism, which we adopt throughout this work, is ordinary differential equations (ODEs) with initial condition:
$ {d\vx}/{dt} = f(\vx, t), \quad \vx(t_0) = \vx_0 $,
where $\vx \in \sR^n$ is the state vector (also called level or stock variables), $f: \sR^n \times \sR \to \sR^n$ the vector field, and $\vx_0 = \begin{bmatrix} x_{0,1}, & \ldots, & x_{0,n} \end{bmatrix}^\top$ the initial values at $t_0$, the initial time.
As a result, when $f$ follows a known, simple form, analytical techniques for ODEs~\citep{walter2013ordinary}, such as the Laplace transform~\citep{mclachlan2014laplace}, can be applied to complex systems analysis.

However, in practice, complex systems are more commonly analyzed through numerical approaches like the first-order Euler method~\citep{ogata2004system} due to the complexity of $f$ when nonlinear feedbacks are present.
This approach enables flexible modeling by requiring only that $f$ be computable, which is crucial since many complex systems admit no analytical solution.
Concretely, the first-order Euler method discretizes the system with time step $\Delta t$ as:
$\vx_{t + \Delta t} = \vx_{t} + \Delta t \cdot f(\vx_{t}, t), \quad \vx_0 = \vx(0)$.
Note that $\vx$ and its discretized counterparts $\vx_t$ and $\vx_{t+\Delta t}$ represent interpretable, meaningful quantities such as populations and resources, rather than abstract hidden states as in RNNs, despite the similarity in their formulations.
Historically, communities studying complex systems have benefited from specialized modeling languages like DYNAMO and STELLA~\citep{radzicki2008origin,richmond1985stella}. 
However, these tools suffer from outdated syntax or proprietary visual interfaces~\citep{forrester2011mit} that limit their usability with LLMs.

\begin{figure*}[t]
  \figremovespace
  \begin{center}
    \includegraphics[width=0.75\textwidth,trim=0 220pt 0 0,clip]{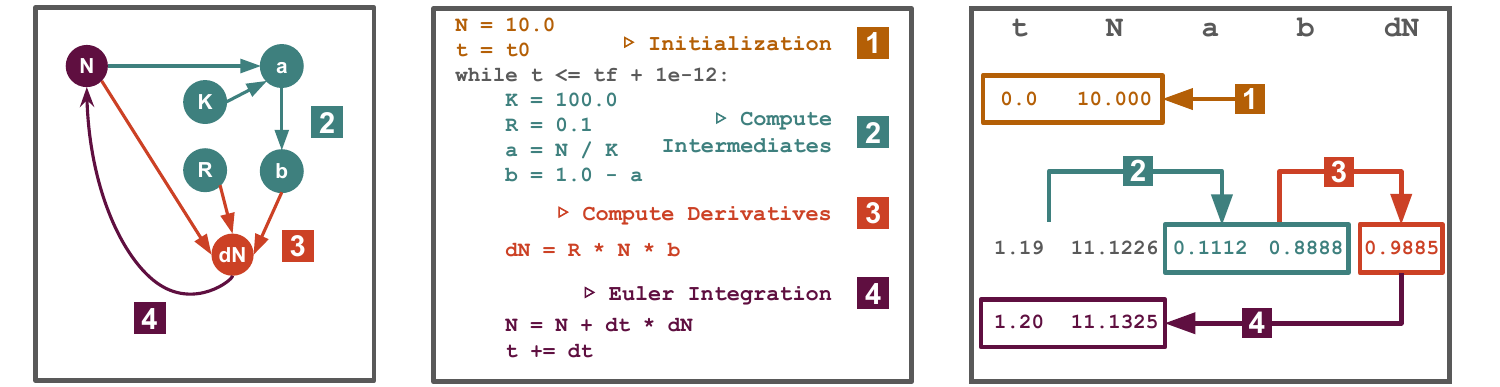}
    \caption{Python representation of a complex system. From left to right: graph representation (for reference), Python code, and computation process. Four essential components are shown: (1) initialization of values ($\vx_0 = \vx(0)$), a loop of (2) intermediate computations, (3) derivative evaluations (i.e., computing $f(\vx_{t}, t)$), and (4) Euler-method integration (i.e., computing $\vx_{t + \Delta t}$ from $\vx_{t}$ and $f(\vx_{t}, t)$). This Python-based representation allows users to easily code and inspect, while leveraging LLM's coding capabilities.}
    \label{figure:code-example}
  \end{center}
\figremovespace
\end{figure*}

\subsection{Unified Representation}

To bridge complex systems with LLMs, we propose a representation of complex systems based on a restricted subset of Python programs, unifying the representations traditionally used in DYNAMO and STELLA.
Python is widely familiar, and LLMs have strong Python coding capabilities.
This avoids creating a new domain-specific language, and enables flexible implementation of $f(\vx, t)$ through general-purpose programming.
We show an example of this representation in Figure~\ref{figure:code-example}.

Besides plain code, we enrich it with documentation and predefined, domain-specific mathematical primitives that aid LLMs in using the representation.
Doing so helps LLMs understand the complex system's semantics, manipulate system dynamics through semantic constructs rather than low-level implementation details, and know what restrictions must be enforced.

\subsection{Proposed Method}

We propose \method, an autonomous method that combines MCTS, LLM, and the unified representation above to discover complex systems that meet a given goal.
We present the overall architecture and components of \method in Figure~\ref{figure:mcts-diagram} and provide details as follows.

\paragraph{Overview and Formalization. }
\method optimizes complex systems by editing them towards a goal $G$, which is a natural-language description of the desired system behavior (for example, ``\texttt{grow population stably}'').
A system $P$ is represented using the restricted subset of Python programs.
The MCTS iteratively builds $T = (V, E)$ with node set $V$ and edge set $E$, each time expanding a selected node $v$ and adding new child node $u$.
We denote the expansion strategy used to generate $u$ as $s_u$, and $u$'s parent as $p_u \triangleq v$ (the root node $0 \in V$ has no parent $p_0$ or expansion strategy $s_0$).
Each node $u \in V$ stores a system $P_u$, an execution record $R_u$, a score $S_u$, and a textual analysis $A_u$, which are produced in the following way:
An LLM Editor produces a new program $P_u$ by patching $P_v$ (e.g. modifying equations, changing feedback links).
$P_u$ is then executed to obtain a new execution record $R_u \leftarrow \textsc{Run}(P_u)$ which includes values of all variables at each timestamp.
The LLM Judge evaluates the quality of system $P_u$ against the specified goal $G$, producing $(A_u, S_u)$: $S_u \in \mathbb{R}$ is a bounded numerical score reflecting how well $P_u$ satisfies goal $G$, and $A_u$ is a textual analysis providing qualitative feedback about the system's behavior and potential improvements.

\begin{figure*}[t]
  \figremovespace
  \begin{center}
    \includegraphics[width=0.81\textwidth,trim=0 0 0 0,clip]{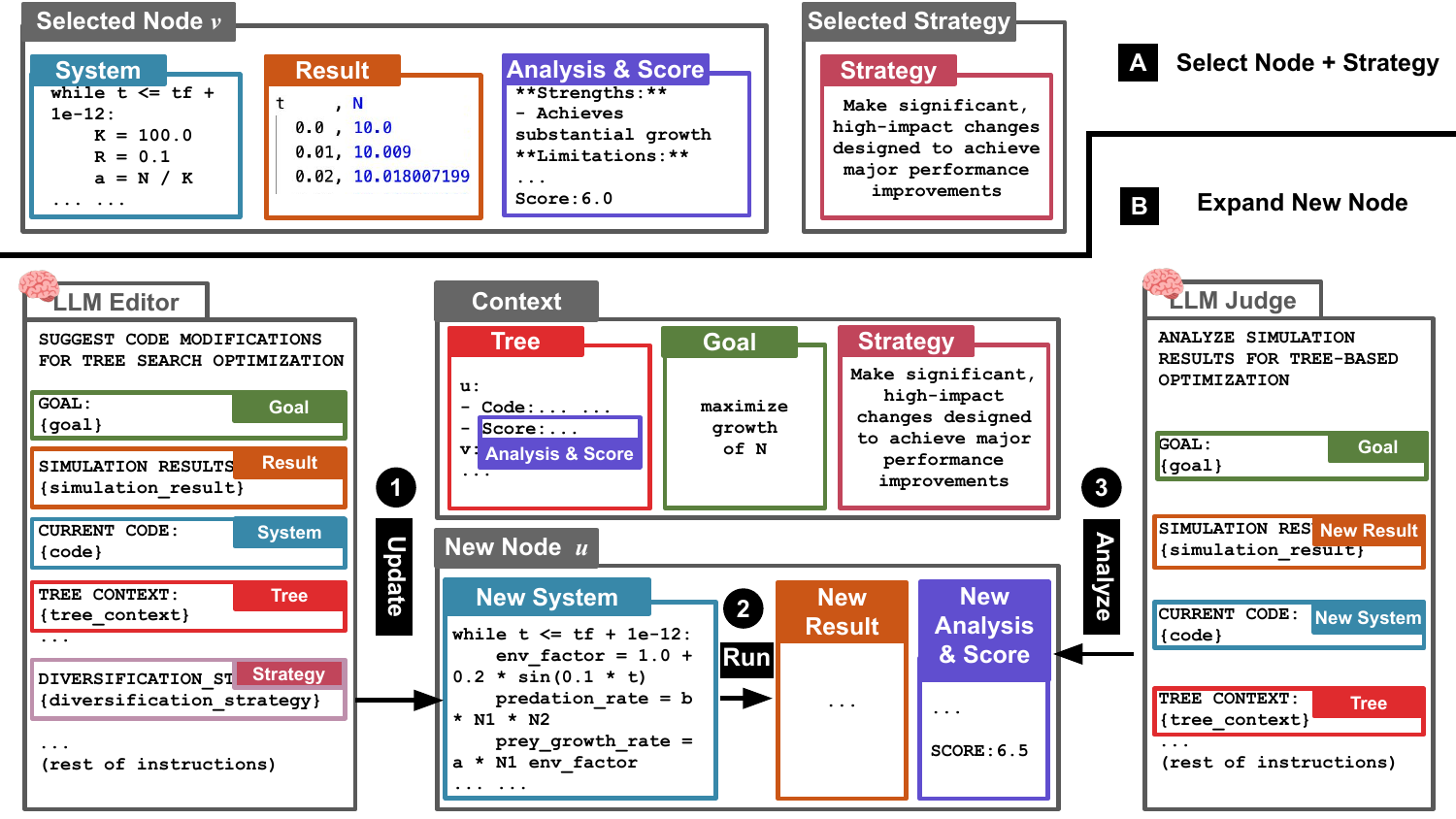}
    \caption{The node expansion process using LLM Editor and Judge. Given a selected node $v$ and expansion strategy $s$, the LLM Editor generates a new system variant $P_u$ by leveraging the parent system $P_v$, its analysis $A_v$, strategy $s$, and goal $G$. The generated system is executed to obtain record $R_u$, then evaluated by the LLM Judge to produce analysis-score pair $(A_u, S_u)$. We embed analysis and scores of all nodes, including the currently selected node, in the tree context to provide the LLM with a comprehensive view of the search tree.}
    \label{figure:llm-judge-and-editor}
  \end{center}
  \figremovespace
\end{figure*}

\paragraph{Initialization. }
\method{} begins by initializing the MCTS tree $T$ with a single root node $v_0$, associated with an initial complex system $P_0$ which can be either provided by the user or generated from a basic template, $P_0$'s execution record $R_0 \leftarrow \textsc{Run}(P_0)$ and the evaluation of $P_0$: $(A_0, S_0) \leftarrow \textsc{LlmJudge}(P_0, R_0, G)$.

\paragraph{Node Selection. }
The node selection process jointly selects $(v, s_u)$, a node $v \in V$ to expand from and an expansion strategy $s_u$ to use, for generating new node $u$ later.
Selection of expansion strategy $s_u$ is simply done by uniformly sampling $s_u \sim \text{Uniform}(\mathcal{S})$ where $\mathcal{S} = \{s_1, s_2, \ldots, s_k\}$ is a fixed set of expansion strategies. 
Selection of $v$ is based on $\textsc{Score}(v)$ that combines the node's performance score $S_v$, its depth $\textsc{Depth}(v)$ in the tree, its expansion count $c_v \in \mathbb{N}$, and its parent $p$'s expansion count $c_p$ (we omit the subscript $v$ and write $p=p_v$):
\method uses the generalized UCT score $\textsc{Score}(v) = S_v + \phi\!\left(c_v, c_p, \tau\right)$.
Here, we have $\phi(\cdot) = -\infty$ if $c_v \ge \tau$ and $ \alpha \cdot \sqrt{\ln(c_p+1)/(c_v+1)} + \gamma \cdot \textsc{Depth}(v)$ otherwise, where $\alpha$ adjusts the amount of exploration, $\gamma > 0$ controls the preference for deeper nodes, and $\tau$ controls the expansion count.
Inspired by \citet{inoue2025wider} and unlike vanilla MCTS, we limit expandable nodes to a subset $V_{\text{exp}} \subseteq V$ and allow repeated expansion of internal nodes until their expansion count $c_v$ reaches a threshold $\tau \in \mathbb{N}$, so a node $v$ is expandable if and only if $v \in V_{\text{exp}}$ and $c_v < \tau$. 
The process selects $v= \arg\max_{v \in V_{\text{exp}}} \textsc{Score}(v)$.

\paragraph{Node Expansion with LLMs. } 
From a selected node $v$ and expansion strategy $s_u$, the LLM Editor generates a new system variant $P_u \leftarrow \textsc{LlmEditor}(P_v, A_v, s, G)$.
Using the system $P_v$, its associated analysis $A_v$, the chosen strategy $s$, and the goal specification $G$, this expansion produces a revised system that satisfies syntactic constraints and semantic requirements while being more aligned with the goal.
Because the system is represented as code, the Editor can make structural edits to $P_v$ rather than only tuning coefficients: it may add or remove state variables, introduce or rewire feedback relations, and change the equations.
The newly generated system $P_u$ is executed to obtain its record $R_u \leftarrow \textsc{Run}(P_u)$, followed by the evaluation  $(A_u, S_u)$ $\leftarrow$ $\textsc{LlmJudge}(P_u, R_u, G)$ that produces a score $S_u$ and a qualitative analysis $A_u$.

\paragraph{Summary. } 
This iterative process of selection, expansion, execution, and evaluation enables progressive improvement toward the target objectives by exploring the space of complex systems.
We formalize the complete search procedure in Algorithm~\ref{algo:mcts} and illustrate particularly the expansion process in Figure~\ref{figure:llm-judge-and-editor}.
Both \textsc{LlmEditor} and \textsc{LlmJudge} are provided with well-crafted prompts covering system context, tree context, and structural constraints, which alongside code-based representation are essential for the method to work.

\subsection{Theoretical Connection}
While designed empirically, \method{} is a principled MCTS variant.
Node selection is a generalized UCT combining progressive widening~\citep{chaslot2008progressive,inoue2025wider} and depth-based bonuses~\citep{blackshaw2025enhancing,wu2025deepsearch}: $\textsc{Score}(v) = S_v + \alpha\sqrt{\ln(c_p{+}1)/(c_v{+}1)} \cdot \mathbb{I}[c_v < \tau] + \gamma \cdot \textsc{Depth}(v)$.
The LLM Editor acts as a stochastic transition kernel over programs, $P_u \sim P_\theta(\cdot \mid P_v, A_v, s, G)$, connecting to learned-transition MCTS and model-based RL~\citep{silver2017mastering,schrittwieser2020mastering}.
The LLM Judge acts as a learned value function providing bounded noisy rewards, connecting to MCTS with biased evaluators~\citep{lisy2013convergence,efroni2019combine}.
That said, we note that the convergence and regret guarantees for MCTS with LLMs remain open problems despite strong empirical performances~\citep{brandfonbrener2024vermcts,katz2024thought}.




\begin{algorithm}[tb!]
  \figremovespace
  \caption{Monte Carlo Tree Search for Complex System Discovery} 
  \label{algo:mcts}
  \begin{center}
    \begin{algorithmic}[1]
    \STATE \textbf{Input:} Initial system $P_0$, goal $G$, iterations $N$, expansion threshold $\tau$
    \STATE \textbf{Output:} Best discovered system $P^*$
    \STATE Initialize: $T = (V, E)$ with root node $0$, $(A_0, S_0) \leftarrow \textsc{LlmJudge}(P_0, \textsc{Run}(P_0), G)$, $P^* \leftarrow P_0$, $S^* \leftarrow S_0$
    \FOR{$i = 1$ to $N$}
        \STATE $v \leftarrow \arg\max_{v \in V_{\text{exp}}} \textsc{Score}(v)$ \hfill $\triangleright$ Select node to expand
        \STATE $s \sim \text{Uniform}(\mathcal{S})$ \hfill $\triangleright$ Sample expansion strategy
        \STATE $P_u \leftarrow \textsc{LlmEditor}(P_v, A_v, s, G)$ \hfill $\triangleright$ Expand
        \STATE $R_u \leftarrow \textsc{Run}(P_u)$ \hfill $\triangleright$ Execute
        \STATE $(A_u, S_u) \leftarrow \textsc{LlmJudge}(P_u, R_u, G)$ \hfill $\triangleright$ Evaluate
        \STATE Add node $u$ to the tree with $(P_u, R_u, A_u, S_u)$
        \STATE $c_v \leftarrow c_v + 1$ \hfill $\triangleright$ Increment expansion count
        \IF{$c_v \geq \tau$}
            \STATE $V_{\text{exp}} \leftarrow V_{\text{exp}} \setminus \{v\}$
            \hfill $\triangleright$ Drop from expandable set
        \ENDIF
        \IF{$S_u > S^*$}
            \STATE $P^* \leftarrow P_u$, $S^* \leftarrow S_u$
        \ENDIF
    \ENDFOR
    \STATE \textbf{return} $P^*$
    \end{algorithmic}
  \end{center}
  \figremovespace
\end{algorithm}

}

{

\section{Experiments}
\label{sec:experiments}

\method enables three capabilities difficult for existing approaches: optimizing for vague natural-language goals, fitting records without complete system skeletons, and interpretable optimization.
We provide qualitative demonstrations and quantitative comparisons for these capabilities.

\subsection{Experimental Details}

\textbf{Data. } We collect complex systems from classical works: \emph{World Dynamics}~\citep{forrester1971world} in the DYNAMO language and 19 systems from the book Modeling Dynamic Biological Systems~\citep{hannon2014modeling} in the STELLA language (20 systems, 20 to 69 integrated variables each), converted into our representation.
We utilize these systems in two ways: (1) as given models to be further optimized towards multiple goals, and (2) as ground truth systems that produce reference records serving as optimization targets.

\textbf{Setup. }
In our experiment, we conduct $20$ node selections, during each of which we expand to $5$ new candidate nodes.
This results in a total of $100$ expansions per search.
We empirically set hyperparameters $\alpha=1$ and $\gamma=2$ for node selection.
The expansion strategy is selected by sampling from a set of candidate strategies listed in Table~\ref{table:expansion-strategies-main}.

\begin{table}[h]
\caption{MCTS expansion strategies.}
\label{table:expansion-strategies-main}
\vskip -0.5em
\begin{footnotesize}
\begin{tabular}{l p{0.62\linewidth}}
\toprule
Strategy & LLM Instructions \\
\midrule
breakthrough & Make significant, high-impact changes for major performance improvements \\
aggressive & Make bold structural or algorithmic changes \\
amplify & Identify and significantly increase the most promising parameters \\
exploratory & Try completely different parameter combinations \\
targeted & Focus on specific high-impact parameters from analysis \\
contrarian & Try approaches opposite to current trends \\
balanced & Make moderate parameter changes with good risk/reward ratio \\
conservative & Make small, incremental parameter adjustments \\
\bottomrule
\end{tabular}
\end{footnotesize}
\end{table}

We use adaptive context subsampling for scalability, keeping the execution record within a context budget $L$ and achieving $O(NL)$ overall complexity.
Because the LLM Editor and Judge define the transition and evaluation operators, we treat their settings as part of the method: all runs use two strong-coding providers, Anthropic Claude Sonnet 4.5 and GPT-5.1, at default decoding settings with structured outputs and up to three retries, and assign crashed or NaN-valued runs a low score so MCTS naturally backtracks. 
Our runs incur moderate computational costs and runtime (\$10--\$50, 30 minutes per run).


\subsection{Fitting an Abstract Goal Described in Natural Language}
\label{section:experiments-fitting-goal}

In this experiment, we task \method with optimizing a given good-enough complex system towards a further ambitious goal. 
The starting system is \emph{World Dynamics}~\citep{forrester1971world}, a published, hand-designed model of the co-evolution of human population, resource utilization, and pollution, containing the most variables in our data and posing challenges for understanding the system dynamics.
Our natural-language goal is to further optimize it to achieve more population growth, less resource usage, and less pollution, all \emph{jointly}:

\begin{tcolorbox}[colback=blue!5!white,colframe=blue!50!white,title={Natural-language goal},breakable,left=0pt,right=0pt,top=0pt,bottom=0pt,boxsep=2pt]
\begin{lstlisting}[style=goal,aboveskip=0pt,belowskip=0pt,xleftmargin=0pt,xrightmargin=0pt]
Balance population, resources, and environment by optimizing to (1) maximize population growth, (2) minimize the resource depletion rate, and (3) minimize the pollution accumulation rate. Seek the best trade-off where the population grows sustainably without depleting resources too quickly or creating excessive pollution. (Important: only change the coefficients in the helper. Do not change any coefficient by more than $50\\%$ to prevent variable explosion.)
\end{lstlisting}
\end{tcolorbox}

Doing so poses intrinsic challenges: the hand-designed system already reaches a kind of Pareto frontier, where optimizing one sub-goal often comes at the expense of others, as the subgoals are conflicting in nature, as Figure~\ref{figure:fitting-single-goal-main} shows.

\begin{figure}[h!]
  \centering
  \includegraphics[width=0.76\columnwidth,trim=0 160pt 400pt 0,clip]{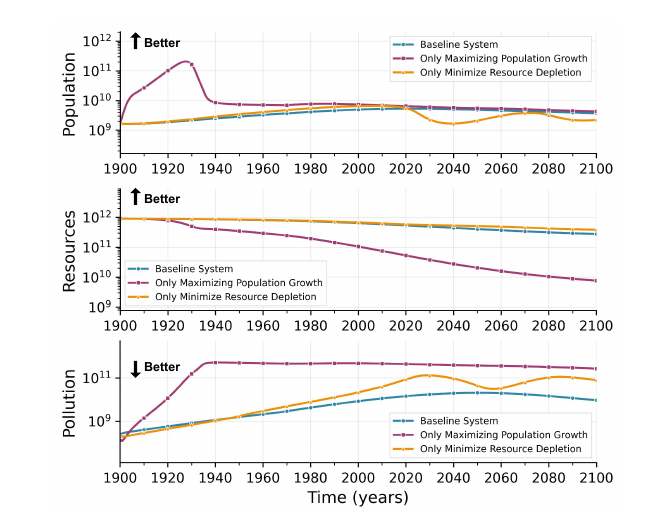}
  \caption{Optimizing for a single goal at a time: focusing on one objective hurts the others, showing competitive subgoals. As in Figure~\ref{figure:fitting-goal}, the Resources panel plots the \emph{remaining} resource stock, so its upward ``better'' direction corresponds to less depletion.}
  \label{figure:fitting-single-goal-main}
\end{figure}

\begin{figure}[ht]
  \figremovespace
  \begin{center}
    \includegraphics[width=0.72\linewidth,trim=0 0 450pt 0,clip]{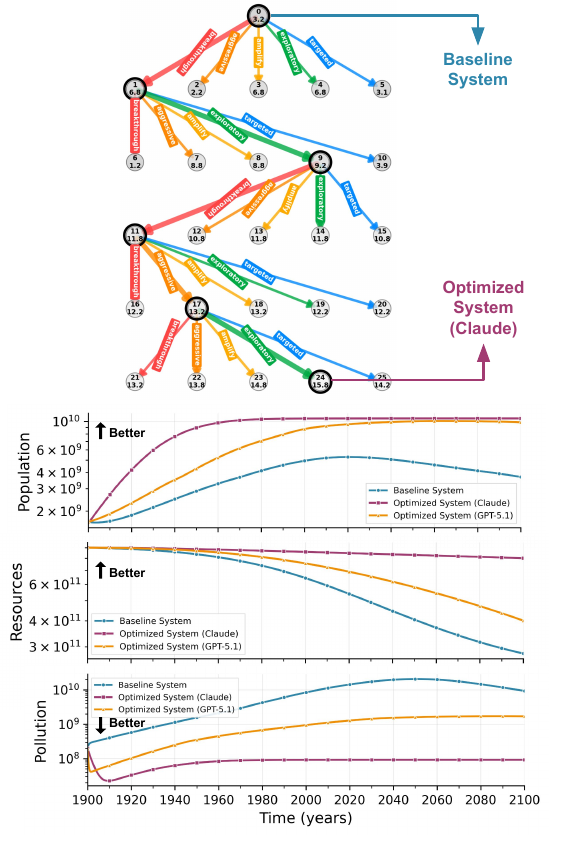}
    \caption{ Illustration of fitting an abstract, language-based goal from \emph{World Dynamics}. Top: The MCTS tree with the path to best node highlighted. Each node $v$ is marked by its index and LLM Judge's score $S_v$. Bottom: The system dynamics for the initial (original) and final complex systems with two LLM backends. We show key variables including Population ($\uparrow$), Pollution ($\downarrow$), and Resources ($\uparrow$). Here Resources denotes the \emph{remaining} natural-resource stock, so a higher value reflects less depletion and is preferable. With \method, the system reaches better states (increased population, less resource depletion, and less pollution).}
    \label{figure:fitting-goal}
  \end{center}
  \figremovespace
\end{figure}

Figure~\ref{figure:fitting-goal} illustrates the MCTS search tree and the resulting system dynamics, and shows how the tree structure guides the search toward higher-scoring solutions.
The resulting system satisfies the vague goal, with all three key variables (population, resource, pollution) improving over the initial system.
Interestingly, with different LLM backends, \method discovers multiple high-performing systems that are structurally distinct, each emphasizing different tradeoffs among objectives (for example, prioritizing population by Claude versus resource by GPT-5.1).

\subsection{Quantitative Studies: Fitting a Concrete Record}
\label{section:experiments-fitting-record}

In this experiment, we task \method with evolving a system (using only a bare-minimum skeleton) to fit the simulated record of a target complex system.
While the strength of \method lies in fitting vague, abstract goals, it can nonetheless be used to fit records from either observations or ground truth systems, by simply loading the record into the goal and running \method as-is.
The ground truth complex system is a population growth model with stochastic components (a stochastic death rate sampled at each time step, introducing volatility intrinsic to the system).
We compare \method with existing approaches based on Optuna~\citep{akiba2019optuna} black-box optimization.
Since the search space for black-box optimization consists of only coefficients, we explore three levels of formula support: no formulae, a simple population-dependent death rate, or the full ground truth formulae, the last of which gives the baseline a serious boost by reducing the task to parameter fitting.
We note this comparison deliberately favors the baseline: each Optuna variant is granted $100$ trials and, in its strongest setting, the full ground-truth formulae, which is strictly more prior structure than \method receives, as \method starts from a bare-minimum skeleton.
To measure the accuracy of fitting a concrete record, we employ both L1 and Dynamic Time Warping (DTW)~\citep{sakoe1978dynamic} distances as metrics. DTW measures trajectory similarity under optimal temporal alignment using a Sakoe-Chiba band ($w{=}250$, 7.1\% of the 3500-step sequence).

\begin{table}[ht]
  \figremovespace
  \caption{L1 and DTW distances to the target record (lower is better). The Formulae columns indicate how much of the ground-truth structure each method is given: No (no formulae), S.\ (a simple population-dependent death rate), or F.\ (the full ground-truth formulae). \method, given no formulae, achieves lower error than the Optuna baseline given the full formulae.}
  \label{table:performance-comparison}
  \begin{center}
    \small
    \definecolor{color1}{HTML}{006666}
    \definecolor{color2}{HTML}{AA3366}
    \definecolor{color3}{HTML}{5533AA}
    \definecolor{color4}{HTML}{CC4400}
    \definecolor{color5}{HTML}{00AA44}
    \begin{tabular}{p{2.5cm}p{0.12cm}p{0.12cm}p{0.12cm}rr}
      \toprule
      \multirow{2}{1.5cm}{\centering Method} & \multicolumn{3}{c}{Formulae} & \multirow{2}{*}{\centering L1} & \multirow{2}{*}{\centering DTW} \\
      & No & S. & F. & & \\
      \cmidrule{2-4}
      \midrule
      \textcolor{color1}{Optuna} & \textcolor{color1}{$\checkmark$} & & & \textcolor{color1}{$29.06$} & \textcolor{color1}{$5503.68$} \\
      \textcolor{color2}{Optuna} & & \textcolor{color2}{$\checkmark$} & & \textcolor{color2}{$26.01$} & \textcolor{color2}{$4927.94$} \\
      \textcolor{color3}{Optuna} & & & \textcolor{color3}{$\checkmark$} & \textcolor{color3}{$3.71$} & \textcolor{color3}{${477.52}$} \\
      \textcolor{color4}{\method (Claude)} & \textcolor{color4}{$\checkmark$} & & & \textcolor{color4}{${3.29}$} & \textcolor{color4}{${757.21}$} \\
      \textcolor{color5}{\method (GPT-5.1)} & \textcolor{color5}{$\checkmark$} & & & \textcolor{color5}{$\mathbf{2.22}$} & \textcolor{color5}{$\mathbf{433.13}$} \\
      \bottomrule
    \end{tabular}
  \end{center}
  \figremovespace
\end{table}
\begin{figure}[ht]
  \figremovespace
  \begin{center}
    \includegraphics[width=0.70\linewidth]{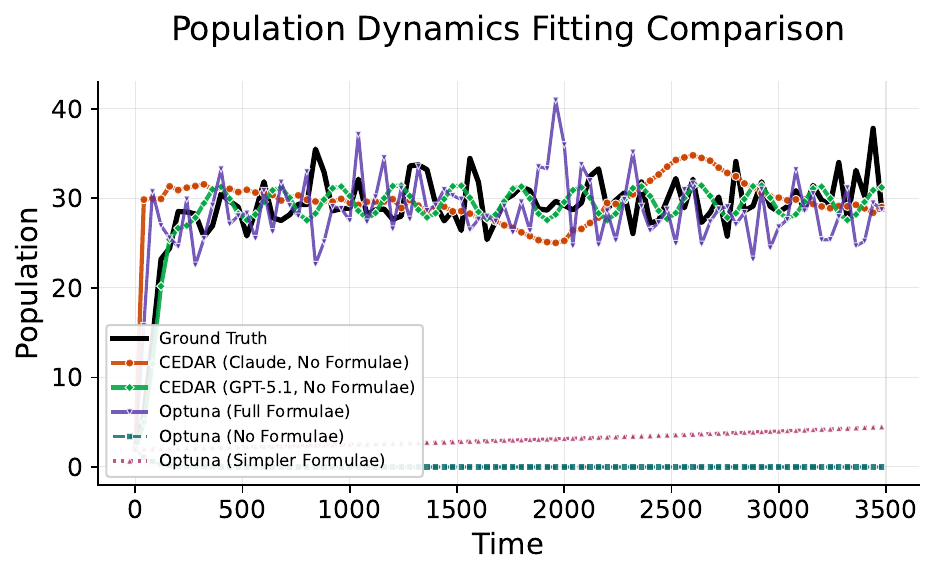}
    \caption{System dynamics (population) when fitting a concrete record. \method, without predefined formulae, tracks the ground-truth trajectory closely, while the no- and simple-formulae Optuna baselines collapse to near-constant populations; only Optuna with the full formulae stays competitive. (All trajectories rise together during the early ramp-up, where the legend overlaps the curves.)}
    \label{figure:fitting-comparison}
  \end{center}
  \figremovespace
\end{figure}

As shown in Table~\ref{table:performance-comparison} and Figure~\ref{figure:fitting-comparison}, the Optuna baselines are severely limited by the skeletal formulae: the no/simple-formulae variants are clearly worse, quantitatively and qualitatively, than the full-formulae variant. Meanwhile, \method, even without predefined formulations, finds solutions that outperform Optuna with a full formula skeleton.
Across 10 seeds, the ground truth's stochasticity (a death rate drawn from a normal distribution each step, scaled by population) shows volatility is intrinsic to the system, not an optimizer artifact; \method{}'s record generally falls within the ground-truth variability and captures the magnitude and timing of major peaks and dips.
Furthermore, even with full-formulae support and 100 trials, Optuna does not perfectly match the dynamics (L1 $3.71$--$4.26$ across two runs), as the stochastic noise floor and high-dimensional nonlinear parameter space make purely numeric search insufficient.

\subsection{Interpretability}

For the \emph{World Dynamics} system above, we examine the interpretability of \method through the LLM's analysis of candidate systems, highlighting in Table~\ref{table:llm-responses} the LLM responses along the path that produces the best system in the search tree of Figure~\ref{figure:fitting-goal}.
\method exposes clear, human-readable analyses at each step that document how the three subgoals are balanced.
The LLM Judge analyzes issues with balanced focus on subgoals, while the LLM Editor first evaluates overall performance, then fine-tunes individual subgoals, also attending to unmentioned aspects (e.g., capital productivity).
These analyses are the stated rationale accompanying each edit, aiding inspection of the search; they are not a verified causal account of how structure produces emergent behavior, which remains an open problem.

\begin{table}[ht]
  \figremovespace
  \caption{ LLM Editor and Judge responses along the best path in the tree. We show the LLM-summarized version with manually colored \textcolor{populationcolor}{Population}, \textcolor{pollutioncolor}{pollution}, and \textcolor{resourcecolor}{resources}.}
  \label{table:llm-responses}
  \begin{center}
    \tiny
    \emergencystretch=1em
    \sloppy
    \begin{tabular}{@{}>{\raggedright\arraybackslash}p{4.1cm}@{\hskip 4pt}>{\raggedright\arraybackslash}p{4.0cm}@{}}
      \toprule
      \textbf{LLM Editor Response} & \textbf{LLM Judge Response} \\
      \midrule
      (No editor response for root, which is the initial system given as is.) & \texttt{Unsustainable overshoot behavior - \textcolor{resourcecolor}{69\% resource depletion}, \textcolor{pollutioncolor}{46x pollution increase}} \\
      \midrule
      \texttt{Breakthrough approach - 25-40\% efficiency improvements targeting 6-8 score} & \texttt{Sustainable progress - \textcolor{populationcolor}{population to 5.46B}, \textcolor{resourcecolor}{30\% resources remaining}, controlled \textcolor{pollutioncolor}{pollution}} \\
      \midrule
      \texttt{Exploratory approach - radical efficiency through enhanced capital productivity} & \texttt{Paradigm shift - \textcolor{populationcolor}{7.0B population}, \textcolor{resourcecolor}{78\% resource conservation}, \textcolor{pollutioncolor}{near-zero pollution}} \\
      \midrule
      \texttt{Ultra-aggressive conservation - pushing toward exceptional 9.5+ performance} & \texttt{Breakthrough sustainability - \textcolor{populationcolor}{3.2x population}, \textcolor{resourcecolor}{36\% resource depletion}, \textcolor{pollutioncolor}{90\% pollution improvement}} \\
      \midrule
      \texttt{Revolutionary breakthrough - \textcolor{resourcecolor}{75\% resource reduction}, \textcolor{pollutioncolor}{90\% pollution control}, 300\% capital productivity} & \texttt{Best-in-tree performance - \textcolor{populationcolor}{6.3x population}, \textcolor{resourcecolor}{59\% resource conservation}, exceptional \textcolor{pollutioncolor}{pollution control}} \\
      \midrule
      \texttt{Extreme efficiency breakthrough - \textcolor{resourcecolor}{95\% resource conservation}, \textcolor{pollutioncolor}{98\% pollution reduction}, 15x capital productivity} & \texttt{Holy grail achievement - \textcolor{populationcolor}{6.4x population growth} with only \textcolor{resourcecolor}{19.9\% resource depletion}} \\
      \bottomrule
    \end{tabular}
  \end{center}
  \figremovespace
\end{table}

\subsection{Why Use MCTS with LLMs: Diversity of Solutions and Better Performance}

We find the benefits of MCTS to be threefold.
\textbf{First, MCTS helps preserve solution diversity.}
For the record-fitting experiment, we sample nodes within a single run whose scores are close to the best solution and visualize their systems' populations in Figure~\ref{figure:diversity}. All satisfy the goal, yet follow distinct trajectories: unlike the Optuna baselines, \method avoids collapsing to a single overfitting-like point, which enables better decision-making through sensitivity analysis.

\begin{figure}[ht]
  \figremovespace
  \begin{center}
    \includegraphics[width=0.90\linewidth]{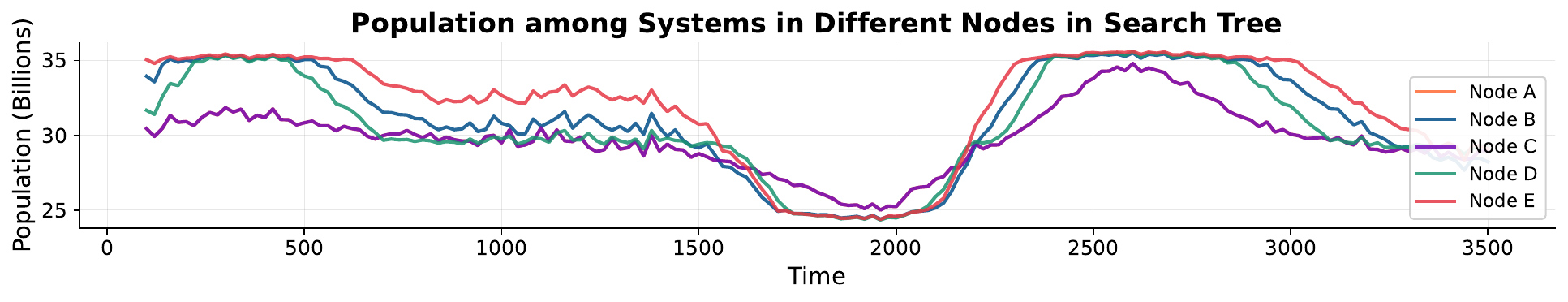}
    \caption{Diversity of solutions in the search tree showing different population dynamics after systems pass the initial phase ($t>100$).}
    \label{figure:diversity}
  \end{center}
  \figremovespace
\end{figure}

\textbf{Second, MCTS leads to better performance.}
Our ablation against linear search (Figure~\ref{figure:linear-vs-mcts}) shows MCTS yields higher-scoring nodes and smoother trajectories that avoid overfitting-like behavior, indicating MCTS with LLMs is crucial for performance gains.

\begin{figure}[h!]
  \figremovespace
  \begin{center}
    \begin{minipage}[t]{0.29\columnwidth}
    \centering
    \includegraphics[width=\linewidth]{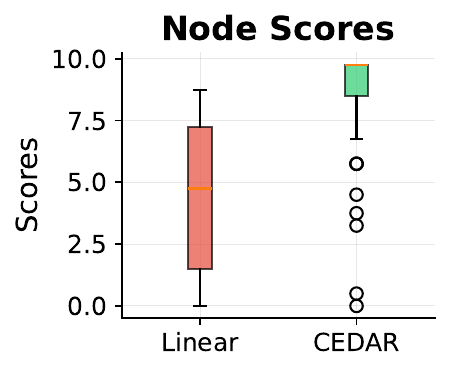}
    \end{minipage}
    \hfill
    \begin{minipage}[t]{0.58\columnwidth}
    \centering
    \includegraphics[width=\linewidth]{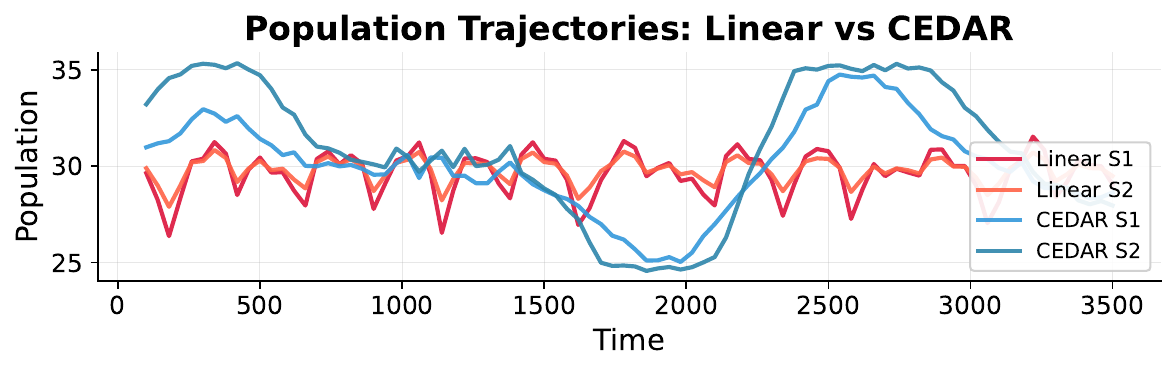}
    \end{minipage}
    \caption{MCTS versus linear search. Left: MCTS attains higher-scoring nodes than linear search. Right: the resulting population dynamics are smoother and avoid the overfitting-like behavior of linear search, showing that tree-based exploration drives \method's performance.}
    \label{figure:linear-vs-mcts}
  \end{center}
  \figremovespace
\end{figure}

\textbf{Third, MCTS supports open-ended discovery of structurally distinct solutions.}
Because MCTS builds a branching tree rather than converging to a single point, it preserves diversity, which matters in ALife contexts where exploring a range of emergent behaviors is valuable alongside optimizing one objective. This diveristy leads to Ddistinct high-performing systems, each emphasizing different tradeoffs among objectives.
}

\section{Conclusion}
We propose \method, which casts complex system discovery as evolutionary search~\citep{nisioti2024text,lehman2023evolution} guided by MCTS: an LLM Editor acts as a variation operator and an LLM Judge as a fitness function.
Empirically, on classical complex systems from social and biological modeling, \method optimizes challenging multi-objective goals, outperforms black-box optimization without predefined formulas, and surfaces interpretable, step-by-step analyses of how the Editor and Judge reshape the system.
Together, these capabilities open new directions for studying and exploring emergent phenomena in artificial life.
Several limitations point to future work. The LLM Judge both scores candidates and shares a model class with the Editor, risking circularity, as the abstract-goal setting still relies on LLM judgment. We report trends rather than tight statistical claims, and the in-depth experiments focus on two systems. 
A natural next step is systematic benchmarking, along with classical evolutionary and multi-objective optimization methods, to delineate when LLM-based variation and evaluation are necessary rather than helpful.

\newpage
\footnotesize
\bibliographystyle{apalike}  
\bibliography{ref}

\clearpage
\onecolumn 
\appendix
\onecolumn 
\appendix
\clearpage
{\Large APPENDIX}
{

\section*{Appendix Overview}
\label{appendix:overview}

The appendix is organized as follows.
Since appendix references are not included in the main submission text, this overview serves as a guide to locating supplementary material.

\paragraph{Python Implementation of Complex Systems (Section~\ref{appendix:code-core-world-dynamics}).}
Details the unified Python-based representation, including a minimal code example, the full \emph{World Dynamics} system code, and the documentation and domain-specific math primitives provided to LLMs.

\paragraph{Method Details (Section~\ref{appendix:method-details}).}
Contains the full prompt templates for LLM Judge and LLM Editor (Section~\ref{appendix:llm-contextualization}), and the original node selection hyperparameter rationale (Section~\ref{appendix:original-node-selection}).

\paragraph{Theoretical Connection: Full Details (Section~\ref{appendix:theoretical-connection}).}
Presents the full derivations of node selection as a UCT generalization, LLM Editor as a learned transition kernel, and LLM Judge as a learned value function, with equations and discussion of open theoretical problems.

\paragraph{Experiment Details (Section~\ref{appendix:experiment-details}).}
Contains dataset variable statistics (Section~\ref{appendix:data-statistics}), the full expansion strategy set (Section~\ref{appendix:expansion-strategies}), engineering decisions for scalability and stability (Section~\ref{appendix:scalability-and-stability}), and computational cost and runtime (Section~\ref{appendix:computation-cost-and-runtime}).

\paragraph{Fitting an Abstract Goal: Details (Section~\ref{appendix:experiments-fitting-goal-details}).}
Contains the \emph{World Dynamics} system structure and variable definitions (Section~\ref{appendix:world-0dynamics-complex-system-details}), the exact natural-language goal text (Section~\ref{appendix:natural-language-goal-for-optimizing-complex-systems}), and a quantitative demonstration of competing-subgoal challenges (Section~\ref{appendix:intrinsic-challenges-of-optimizing-complex-systems-with-competing-subgoals}).

\paragraph{Fitting a Concrete Record: Details (Section~\ref{appendix:experiments-fitting-record-details}).}
Contains the ground truth system implementation (Section~\ref{appendix:experiments-fitting-record-details-ground-truth-system}), the three levels of Optuna baseline code (Section~\ref{appendix:experiments-fitting-record-details-baseline-systems}), the DTW metric definition (Section~\ref{dynamic-time-warping-distance}), stochasticity analysis across seeds (Section~\ref{appendix:stochasticity-in-ground-truth-complex-system}), and an additional Optuna run (Section~\ref{appendix:optunas-performance-on-baseline-with-full-formulae}).

\paragraph{Interpretability Analysis (Section~\ref{appendix:interpretability-details}).}
Contains full LLM Judge and Editor transcripts from the \emph{World Dynamics} optimization run.

}

{

\section{Python Implementation of Complex Systems}
\label{appendix:code-core-world-dynamics}

Here we detail our proposed unified representation of complex systems based on a restricted subset of Python programs.
Concretely, the components are divided into four parts:
(1) initialization, a loop of (2) intermediate computations, (3) derivative evaluations, and (4) Euler-method integration.
We show a minimal example of a program with these components below:

\begin{lstlisting}
# (1) initialization 
dt = 0.01
t0 = 0.0
tf = 120.0
K = 100.0
R = 0.1
N = 10.0

t = t0
while t <= tf + 1e-12:

    # (2) intermediate computations
    carrying_capacity_factor = 1.0 - N / K
    growth_rate = R * N * carrying_capacity_factor

    # (3) derivative evaluations
    dN = growth_rate  # dN = R * N * (1 - N/K)

    # (4) Euler integration
    N = N + dt * dN
    t += dt
\end{lstlisting}

In practice, we find that including extensive comments in the program is not only helpful for humans to understand, but also beneficial for LLMs. 
Furthermore, to help LLMs' tool use, we extend the code with further documentation and predefined math primitives, such as common math functions, delay, and smoothing functions that are frequently used in complex systems.
The benefits are twofold:
(1) Such comments and documents help LLMs to understand the complex system's semantics for better optimization, and the standardized functions ease LLMs' job in generating code for new systems, and 
(2) more importantly, it instructs LLMs on what restrictions should be enforced, which is important since several critical restrictions, such as preserving the Euler-method integration, must be respected for the program to execute successfully.

We show our representation of the complex system of the \emph{World Dynamics} system~\citep{forrester1971world}. 
As it is the one with the largest number of variables in our dataset, this code below demonstrates the full-fledged form of our representation, including how we enforce the constraints for LLMs and providing semantics to help the optimization.
We show the code in two parts: First, we show the code that directly describes the complex system's dynamics:
\begin{lstlisting}
# ===================== BEGIN CONFIG =========================
# Simulation configuration parameters (DO NOT CHANGE - these are set by the system)
dt = 0.2  # time step size
t0 = 1900.0  # start time
tf = 2100.0  # end time
seed = 1234  # random seed (set None for nondeterministic runs)
# ====================== END CONFIG ==========================

# ====================== BEGIN STATE =========================
# STATE VARIABLES - integrated over time (must define derivatives below)
P = 1.65e9  # population (people) # NECESSARY
NR = 900e9  # natural resources (natural resource units) # NECESSARY
CI = 0.4e9  # capital investment (capital units) # NECESSARY
POL = 0.2e9  # pollution (pollution units) # NECESSARY
CIAF = 0.2  # capital-investment-in-agriculture fraction (dimensionless) # NECESSARY
# ======================= END STATE ==========================

# ================ BEGIN ALGEBRAIC (OPTIONAL) ================
# Variables are now declared in the HELPERS section within the simulation loop

# ====================== BEGIN PARAMS ========================
# CONSTANT PARAMETERS used in equations
# (All constants are inlined in the code)
# ======================= END PARAMS =========================

# ======================= MAIN LOOP ==========================
t = t0
while t <= tf + 1e-12:
    # =================== BEGIN INPUTS_T =====================
    # TIME-DEPENDENT INPUTS - use wrappers like sin(), exp(), graph()
    # (No time-dependent inputs in this model)
    # ==================== END INPUTS_T ======================

    # ================= BEGIN INPUTS_RND =====================
    # RANDOM INPUTS - use gauss(), uniform() wrappers (no STATE usage).
    # (No random inputs in this model)
    # ================== END INPUTS_RND ======================

    # =================== BEGIN HELPERS ======================
    # HELPERS - All computed variables (algebraic and intermediate expressions)
    # in dependency order.

    # Basic ratios and derived quantities
    CIR = CI / P  # capital-investment ratio (capital units/person)
    CR = P / (135e6 * 26.5)  # crowding ratio (dimensionless)
    NRFR = NR / 900e9  # natural-resource fraction remaining (dimensionless)
    POLR = POL / 3.6e9  # pollution ratio (dimensionless)

    # Material standard of living components
    CIAF_current = CIAF  # current capital-investment-in-agriculture fraction
    NREM = graph(
        NRFR, ((0, 0), (0.25, 0.15), (0.5, 0.5), (0.75, 0.85), (1, 1))
    )  # natural-resource-extraction multiplier
    ECIR = CIR * (1 - CIAF_current) * NREM / (1 - 0.3)  # effective-capital-investment ratio
    MSL = ECIR / 1  # material standard of living (dimensionless)

    # Birth rate multipliers
    BRMM = graph(
        MSL, ((0, 1.2), (1, 1), (2, 0.85), (3, 0.75), (4, 0.7), (5, 0.7))
    )  # birth-rate-from-material multiplier
    BRCM = graph(
        CR, ((0, 1.05), (1, 1), (2, 0.9), (3, 0.7), (4, 0.6), (5, 0.55))
    )  # birth-rate-from-crowding multiplier
    BRPM = graph(
        POLR, ((0, 1.02), (10, 0.9), (20, 0.7), (30, 0.4), (40, 0.25), (50, 0.15), (60, 0.1))
    )  # birth-rate-from-pollution multiplier

    # Food ratio components
    CIRA = CIR * CIAF_current / 0.3  # capital-investment ratio in agriculture
    FPCI = graph(
        CIRA, ((0, 0.5), (1, 1), (2, 1.4), (3, 1.7), (4, 1.9), (5, 2.05), (6, 2.2))
    )  # food potential from capital investment
    FCM = graph(CR, ((0, 2.4), (1, 1), (2, 0.6), (3, 0.4), (4, 0.3), (5, 0.2)))  # food-from-crowding multiplier
    FPM = graph(
        POLR, ((0, 1.02), (10, 0.9), (20, 0.65), (30, 0.35), (40, 0.2), (50, 0.1), (60, 0.05))
    )  # food-from-pollution multiplier
    FR = FPCI * FCM * FPM * (1 if t >= 1970 else 1) / 1  # food ratio
    BRFM = graph(FR, ((0, 0), (1, 1), (2, 1.6), (3, 1.9), (4, 2)))  # birth-rate-from-food multiplier

    # Death rate multipliers
    DRMM = graph(
        MSL,
        (
            (0, 3),
            (0.5, 1.8),
            (1, 1),
            (1.5, 0.8),
            (2, 0.7),
            (2.5, 0.6),
            (3, 0.53),
            (3.5, 0.5),
            (4, 0.5),
            (4.5, 0.5),
            (5, 0.5),
        ),
    )  # death-rate-from-material multiplier
    DRCM = graph(CR, ((0, 0.9), (1, 1), (2, 1.2), (3, 1.5), (4, 1.9), (5, 3)))  # death-rate-from-crowding multiplier
    DRPM = graph(
        POLR, ((0, 0.92), (10, 1.3), (20, 2), (30, 3.2), (40, 4.8), (50, 6.8), (60, 9.2))
    )  # death-rate-from-pollution multiplier
    DRFM = graph(
        FR, ((0, 30), (0.25, 3), (0.5, 2), (0.75, 1.4), (1, 1), (1.25, 0.7), (1.5, 0.6), (1.75, 0.5), (2, 0.5))
    )  # death-rate-from-food multiplier

    # Natural resource usage multiplier
    NRMM = graph(
        MSL, ((0, 0), (1, 1), (2, 1.8), (3, 2.4), (4, 2.9), (5, 3.3), (6, 3.6), (7, 3.8), (8, 3.9), (9, 3.95), (10, 4))
    )  # natural-resource-from-material multiplier

    # Capital investment multiplier
    CIM = graph(MSL, ((0, 0.1), (1, 1.0), (2, 1.8), (3, 2.4), (4, 2.8), (5, 3)))  # capital-investment multiplier

    # Pollution components
    POLCM = graph(CIR, ((0, 0.05), (1, 1), (2, 3), (3, 5.4), (4, 7.4), (5, 8)))  # pollution-from-capital multiplier
    POLAT = graph(
        POLR, ((0, 0.6), (10, 2.5), (20, 5), (30, 8), (40, 11.5), (50, 15.5), (60, 20))
    )  # pollution-absorption time

    # Quality of life components
    QLM = graph(MSL, ((0, 0.2), (1, 1), (2, 1.7), (3, 2.3), (4, 2.7), (5, 2.9)))  # quality of life from material
    QLC = graph(
        CR,
        (
            (0, 2),
            (0.5, 1.3),
            (1, 1),
            (1.5, 0.75),
            (2, 0.55),
            (2.5, 0.45),
            (3, 0.38),
            (3.5, 0.3),
            (4, 0.25),
            (4.5, 0.22),
            (5, 0.2),
        ),
    )  # quality of life from crowding
    QLF = graph(FR, ((0, 0), (1, 1), (2, 1.8), (3, 2.4), (4, 2.7)))  # quality of life from food
    QLP = graph(
        POLR, ((0, 1.04), (10, 0.85), (20, 0.6), (30, 0.3), (40, 0.15), (50, 0.05), (60, 0.02))
    )  # quality of life from pollution
    QL = 1 * QLM * QLC * QLF * QLP  # quality of life

    # Capital investment fraction adjustment components
    CFIFR = graph(FR, ((0, 1), (0.5, 0.6), (1, 0.3), (1.5, 0.15), (2, 0.1)))  # capital fraction indicated by food ratio
    CIQR = graph(QLM / QLF, ((0, 0.7), (0.5, 0.8), (1, 1), (1.5, 1.5), (2, 2)))  # capital-investment-from-quality ratio

    # ==================== END HELPERS =======================

    # ================= BEGIN DERIVATIVES ====================
    # DERIVATIVES for each STATE variable.

    # Birth and death rates
    BR = P * (0.04 if t >= 1970 else 0.04) * BRFM * BRMM * BRCM * BRPM  # birth rate
    DR = P * (0.028 if t >= 1970 else 0.028) * DRMM * DRPM * DRFM * DRCM  # death rate

    # Natural resource usage rate
    NRUR = P * (1 if t >= 1970 else 1) * NRMM  # natural-resource-usage rate

    # Capital investment flows
    CIG = P * CIM * (0.05 if t >= 1970 else 0.05)  # capital-investment generation
    CID = CI * (0.025 if t >= 1970 else 0.025)  # capital-investment discard

    # Pollution flows
    POLG = P * (1 if t >= 1970 else 1) * POLCM  # pollution generation
    POLA = POL / POLAT  # pollution absorption

    # State derivatives
    dP = BR - DR
    dNR = -NRUR
    dCI = CIG - CID
    dPOL = POLG - POLA
    dCIAF = (1 / 15) * (CFIFR * CIQR - CIAF)
    # ================== END DERIVATIVES =====================
    
    # --------------- Euler integration (engine; do not edit) ---------------
    P = P + dt * dP
    NR = NR + dt * dNR
    CI = CI + dt * dCI
    POL = POL + dt * dPOL
    CIAF = CIAF + dt * dCIAF

    # Advance time
    t += dt
\end{lstlisting}

Then, we show the documentation and math functions. The code above is placed between \texttt{BEGIN WHOLE SYSTEM} and \texttt{END WHOLE SYSTEM}.

\begin{lstlisting}
# ============================================================
# sim_base.py - Minimal Euler simulator with edit markers
#
# PURPOSE:
# - Self-contained forward Euler simulator:
#       X(t+dt) = X(t) + dt * dX/dt
#
# HOW TO EDIT:
# - Only change code between the "BEGIN ... / END ..." markers below.
# - Do NOT modify imports, wrappers, loop structure, Euler integration, or OUTPUT section.
# - Keep STATE, ALGEBRAIC, HELPERS, and DERIVATIVES consistent.
# - NEVER change simulation configuration parameters (dt, t0, tf, seed) in the CONFIG section.
#
# VARIABLE MODIFICATION RULES:
# - STATE VARIABLES: You can add new state variables. You can delete state variables ONLY if they 
#   are not marked as "# NECESSARY" in comments. All state variables must have derivatives.
# - ALGEBRAIC VARIABLES: You can add new algebraic variables. You can delete algebraic variables 
#   ONLY if they are not marked as "# NECESSARY" in comments.
# - HELPERS: You can freely add, remove, and modify helper variables. You can change their 
#   computation formulas and reorder them, as long as the code remains valid.
# - When converting from other systems: Mark essential state and algebraic variables with 
#   "# NECESSARY" comments so optimization/editing knows which variables cannot be deleted.
#
# VARIABLE TYPES:
# - STATE: variables integrated over time (must have derivatives).
# - INPUTS_T: pure functions of time (cannot depend on STATE).
# - INPUTS_RND: fresh random values each step (cannot depend on STATE).
# - HELPERS: all computed variables (algebraic and intermediate expressions)
#            in dependency order. Can use wrappers like sin(), exp(), graph(),
#            delay(), smth1() - e.g. nonlinear damping from a table.
# - DERIVATIVES: d/dt for each STATE variable.
#
# TRACE CONTENT (per step, in fixed order):
#   1) t            - current time
#   2) STATE variables
#   3) HELPERS (all computed variables)
#   4) DERIVATIVES
#
# OUTPUT:
# - If --csv <path> is provided -> write full simulation to that CSV file.
# - Otherwise -> print the raw CSV (header + all rows) to stdout.
# - Columns/ordering come from the snapshot dict defined inside the loop.
# ============================================================

import math
import random
import argparse
import pandas as pd  # used only for CSV export
from typing import Iterable, Tuple

# ------------------ Wrapper Functions -----------------------
# These wrappers hide Python semantics; LLMs should ONLY call these.

def sin(x: float) -> float:
    """Sine function (angle in radians)."""
    return math.sin(x)

def cos(x: float) -> float:
    """Cosine function (angle in radians)."""
    return math.cos(x)

def exp(x: float) -> float:
    """Exponential function e^x."""
    return math.exp(x)

def tanh(x: float) -> float:
    """Hyperbolic tangent."""
    return math.tanh(x)

def sqrt(x: float) -> float:
    """Square root."""
    return math.sqrt(x)

def log(x: float) -> float:
    """Natural logarithm."""
    return math.log(x)

def gauss(std: float) -> float:
    """Gaussian random variable with mean=0 and standard deviation=std."""
    return random.gauss(0.0, std)

def uniform(lo: float, hi: float) -> float:
    """Uniform random variable between lo and hi."""
    return random.uniform(lo, hi)

def graph(v: float, table: Iterable[Tuple[float, float]]) -> float:
    """
    Lookup helper with linear interpolation.
    - table is an iterable of (x, y) points sorted by x.
    - If v < x0, return y0.
    - If v > xN, return yN.
    - Else, linearly interpolate between nearest points.
    """
    table = list(table)
    if not table:
        raise ValueError("graph() called with empty table")

    if v <= table[0][0]:
        return table[0][1]
    if v >= table[-1][0]:
        return table[-1][1]

    for i in range(len(table) - 1):
        x0, y0 = table[i]
        x1, y1 = table[i + 1]
        if x0 <= v <= x1:
            if x1 == x0:
                return y0
            frac = (v - x0) / (x1 - x0)
            return y0 + frac * (y1 - y0)
    return table[-1][1]

# ------------------ Delay/Smooth System ---------------------
# STELLA-compatible delay/smooth functions with time-based semantics.
# - All functions require a 'name' parameter for unique identification
# - Updates occur when functions are called during helper computation  
# - Fixed delays use time-indexed history with linear interpolation
# - Smooth functions use stock-based integration with proper dt scaling
# - All functions are dt-independent (same behavior regardless of step size)

_delay_states = {}  # Registry for all delay/smooth function states

def delay(input_val: float, delay_time: float, name: str, initial_value: float = None) -> float:
    """Fixed lag delay - returns input value from delay_time ago."""
    key = f"delay_{name}_{delay_time}"
    
    if key not in _delay_states:
        init_val = input_val if initial_value is None else initial_value
        _delay_states[key] = {
            'history': [(t, init_val)],
            'last_t': t
        }
    
    state = _delay_states[key]
    
    # Add current input to history (only if time advanced)
    if t > state['last_t']:
        state['history'].append((t, input_val))
        state['last_t'] = t
        
        # Clean old history beyond delay time
        cutoff_time = t - delay_time
        state['history'] = [(t_hist, val) for t_hist, val in state['history'] 
                           if t_hist >= cutoff_time]
    
    # Find value at t - delay_time using linear interpolation
    target_time = t - delay_time
    if not state['history'] or target_time <= state['history'][0][0]:
        return state['history'][0][1]
    
    for i in range(len(state['history']) - 1):
        t1, v1 = state['history'][i]
        t2, v2 = state['history'][i + 1]
        if t1 <= target_time <= t2:
            if t2 == t1:
                return v1
            frac = (target_time - t1) / (t2 - t1)
            return v1 + frac * (v2 - v1)
    
    return state['history'][-1][1]

def smth1(input_val: float, smooth_time: float, name: str, initial_value: float = None) -> float:
    """First-order exponential smooth - stock-based smoothing process."""
    key = f"smth1_{name}_{smooth_time}"
    
    if key not in _delay_states:
        init_val = input_val if initial_value is None else initial_value
        _delay_states[key] = {
            'smooth_of_input': init_val,  # The stock being smoothed
            'last_t': t
        }
    
    state = _delay_states[key]
    
    if t > state['last_t']:
        dt_step = t - state['last_t']
        
        # SMTH1 equations from STELLA:
        # Change_in_Smooth = (Input - Smooth_of_Input) / Averaging_Time
        change_in_smooth = (input_val - state['smooth_of_input']) / smooth_time if smooth_time > 0 else 0.0
        
        # Stock integration: Smooth_of_Input = Smooth_of_Input + dt * Change_In_Smooth
        state['smooth_of_input'] += dt_step * change_in_smooth
        state['last_t'] = t
    
    # SMTH1 returns the smoothed stock value
    return state['smooth_of_input']


# ===================== BEGIN WHOLE SYSTEM ===================
#
#
#
# ===================== END WHOLE SYSTEM ===================

# ========================= OUTPUT (FIXED) ===================
# DO NOT EDIT THIS SECTION.
def _parse_args():
    ap = argparse.ArgumentParser(description="Run Euler simulation and export CSV.")
    ap.add_argument("--csv", dest="csv_path", default=None,
                    help="Write results to this CSV path. If omitted, prints CSV to stdout.")
    return ap.parse_args()

if __name__ == "__main__":
    args = _parse_args()
    df = pd.DataFrame(trace)  # column order = insertion order of dict
    if args.csv_path:
        df.to_csv(args.csv_path, index=False, float_format='%.6f')
        print(f"Simulation complete. Results written to {args.csv_path}")
    else:
        print(df.to_csv(index=False, float_format='%.6f'), end="")
\end{lstlisting}

}

{

\section{Method Details: LLM Contextualization, Scalability and Stability}
\label{appendix:method-details}

\subsection{LLM Contextualization}
\label{appendix:llm-contextualization}

In our design, we pair LLM Judge, which performs evaluations of one complex system, with LLM Editor, which produces new modified systems to better achieve the goal.
Besides the enriched code-based representation detailed in Appendix~\ref{appendix:code-core-world-dynamics},
we also need well-crafted prompts to help LLMs to understand the semantics of complex systems and to generate meaningful modifications towards the goal.
In fact, since the code is in the context of the prompt, both the prompt design and code representation can be seen as a coherent \emph{contextualization} of the information.
Also, we find that this contextualization should include the tree context, as our preliminary studies found that without tree context, the expansion process fails to produce meaningful improvements.

In conclusion, the prompt templates for LLM Judge and LLM Editor are listed below respectively:

\begin{tcolorbox}[colback=blue!5!white,colframe=blue!75!black,title=Prompt for \textsc{LlmJudge},breakable]
\begin{lstlisting}[style=prompt]
ANALYZE SIMULATION RESULTS FOR TREE-BASED OPTIMIZATION

GOAL:
(*\colorbox{yellow!30}{\texttt{\{goal\}}}*)

CURRENT CODE:
(*\colorbox{yellow!30}{\texttt{\{code\}}}*)

SIMULATION RESULTS:
(*\colorbox{yellow!30}{\texttt{\{csv\_data\}}}*)

TREE CONTEXT:
(*\colorbox{yellow!30}{\texttt{\{tree\_context\}}}*)


=== For your reference ===

Format for code in tree context:
The tree context shows only the differences in each reference node compared to current code. 
Since the full current code is shown above, diffs only display lines that are different in reference nodes:
- Lines starting with "+" show what the reference node has instead of current code
- "Code identical to current node" means no differences exist

When analyzing this node's performance, consider:
- How well this specific code variant achieves the optimization goal
- Performance relative to ALL other nodes in the reference (for calibrated scoring)
- Whether this represents meaningful progress in the search space

SCORING GUIDELINES:
ABSOLUTE PERFORMANCE SCORING (depth-constrained scale)
Primary Principle: Score based on how well this code achieves the optimization goal, regardless of other nodes in the tree.

SCORING SCALE (use 2 decimal places like 7.25, 12.75):
- 0.00-2.00: Failed progress - no meaningful progress toward the goal
- 2.00-4.00: Partial progress - some improvement but far from achieving the goal  
- 4.00-6.00: Moderate progress - measurable improvement and moving toward the goal
- 6.00-8.00: Good progress - clear advancement with substantial goal achievement
- 8.00-10.00: Excellent progress - meets most requirements of the optimization goal
- 10.00+: VERY GOOD performance - exceeds the goal expectations significantly
- 20.00+: EXCEPTIONAL performance - far surpasses what the goal was asking for

IMPORTANT CONSTRAINT: Maximum possible score is score <= 10.0 + 2.5 * depth. Within this limit, high scores are encouraged when performance truly merits them.

Use the tree context information to ensure your scoring is well-calibrated relative to all explored alternatives.

=== Task ====

Please provide your analysis in this format:

REASONING: [Start with a brief summary of key findings in the first 200 characters, then provide detailed reasoning about how well this code variant achieves the optimization goal]

SCORE: [numerical score based on the scoring guidelines above]
\end{lstlisting}
\end{tcolorbox}

\begin{tcolorbox}[colback=blue!5!white,colframe=blue!75!black,title=Prompt for \textsc{LlmEditor},breakable]
\begin{lstlisting}[style=prompt]
SUGGEST CODE MODIFICATIONS FOR TREE SEARCH OPTIMIZATION

GOAL:
(*\colorbox{yellow!30}{\texttt{\{goal\}}}*)

CURRENT CODE:
(*\colorbox{yellow!30}{\texttt{\{code\}}}*)

SIMULATION RESULTS:
(*\colorbox{yellow!30}{\texttt{\{csv\_data\}}}*)

TREE CONTEXT:
(*\colorbox{yellow!30}{\texttt{\{tree\_context\}}}*)

=== For your reference ===

Format for code in tree context:
The tree context shows only the differences in each reference node compared to current code. 
Since the full current code is shown above, diffs only display lines that are different in reference nodes:
- Lines starting with "+" show what the reference node has instead of current code
- "Code identical to current node" means no differences exist

How to modify the code to better achieve the goal within the tree search context:
Focus on:
- Parameters within the marked BEGIN/END edit sections
- State variables, constants, and simulation configuration  
- Derivative equations and helper expressions
- Table values and time-dependent inputs
- Learning from tree context and reference node outcomes

When you modify the code:
- Faithfully follow the CURRENT CODE. Starting from EVERYTHING in CURRENT CODE, then 
- only modify code between the "BEGIN ... / END ..." markers.
- DO NOT change imports, wrappers, loop structure, or OUTPUT section.

Tree Search Strategy:
- Consider your position in the search tree
- Learn from performance patterns in tree context
- Use diversification strategy from tree context
- Reference successful/failed approaches from other nodes

SCORING GUIDELINES:
ABSOLUTE PERFORMANCE SCORING (depth-constrained scale)
Primary Principle: Score based on how well this code achieves the optimization goal, regardless of other nodes in the tree.

SCORING SCALE (use 2 decimal places like 7.25, 12.75):
- 0.00-2.00: Failed progress - no meaningful progress toward the goal
- 2.00-4.00: Partial progress - some improvement but far from achieving the goal  
- 4.00-6.00: Moderate progress - measurable improvement and moving toward the goal
- 6.00-8.00: Good progress - clear advancement with substantial goal achievement
- 8.00-10.00: Excellent progress - meets most requirements of the optimization goal
- 10.00+: VERY GOOD performance - exceeds the goal expectations significantly
- 20.00+: EXCEPTIONAL performance - far surpasses what the goal was asking for

IMPORTANT CONSTRAINT: Maximum possible score is score <= 10.0 + 2.5 * depth. Within this limit, high scores are encouraged when performance truly merits them.

HOW TO SCORE:
1. First, evaluate how well this code achieves the goal in absolute terms
2. Use the tree context only to calibrate what score ranges mean - don't let it constrain your scoring
3. For very good performance that exceeds expectations, don't hesitate to give high scores
4. When unsure between score ranges, favor the higher score if genuine progress is evident
5. Large score increases (5+ points) are appropriate for significant improvements
6. Always respect the depth constraint: score <= 10.0 + 2.5 * depth

Remember: Judge this code's actual achievement of the goal, not its relative position in the search tree.

=== Task ====

Please provide your response in this format:

REASONING: [Start with a concise description of your main modification strategy in the first 200 characters, then explain detailed reasoning and assess the expected improvement using the scoring guidelines above]

SELF(*\_*)ASSESSED(*\_*)SCORE: [numerical score based on the scoring guidelines above for your expected improvement]

MODIFIED(*\_*)CODE: [Complete modified Python code following all requirements - start from EVERYTHING in CURRENT CODE, then only modify code between BEGIN/END markers]
\end{lstlisting}
\end{tcolorbox}

\subsection{Original Node Selection Rationale}
\label{appendix:original-node-selection}

The choice of hyperparameters $\alpha$ and $\tau$ reflects empirical behavior of LLM-based optimization on complex systems: high-quality solutions often result from several consecutive refinements among a promising branch, so our choice should favour deeper nodes, cap expansion counts, thus avoiding over-exploring shallow nodes with expensive LLM calls.

}

{

\section{Theoretical Connection: Full Details}
\label{appendix:theoretical-connection}

While \method{} components are designed based on empirical considerations, the algorithm is a principled MCTS variant: It has widening-aware and depth-aware UCT selection, LLM-parameterized transition kernel and value function.
We establish the connections as follows, and we also note that for MCTS with LLMs convergence and regret guarantees remain open problems.

\paragraph{Node Selection as a UCT-Generalization.}
Following UCB1 \citep{kocsis2006bandit}, \method uses $\textsc{Score}(v) = S_v + \phi\!\left(c_v, c_p, \tau\right)$,
where $c_v$ and $c_p$ are expansion counts at $v$ and its parent $p$, and $\tau$ is the progressive widening limit.
We combine two MCTS principles: Progressive Widening~\citep{chaslot2008progressive,inoue2025wider} limiting expansion width, and depth-based bonuses favoring deeper search~\citep{blackshaw2025enhancing,wu2025deepsearch}.
This yields:
\begin{equation}
    \phi
    =
    \underbrace{
        \alpha\sqrt{\frac{\ln\!\left(c_p+1\right)}{c_u+1}}
        \cdot \mathbb{I}[c_v < \tau]
        +
        \beta\,\mathbb{I}[c_v \ge \tau]
    }_{
        \text{Progressive Widening}
    }
    +
    \underbrace{
        \gamma \cdot \textsc{Depth}(v)
    }_{
        \text{Depth-based bonus}
    }
\end{equation}
The node selection in \method{} effectively sets $\beta=-\infty$, positioning itself as a simple combination of MCTS variants with non-uniform branching.
The constraint $c_v<\tau$ prevents overexpansion in vast action spaces like code modification and expensive LLM calls.

\paragraph{LLM Editor as a Learned Transition Kernel.}
MCTS assumes a generative model for transitions $u\sim P(\cdot\mid v,a)$ where $a$ denotes an expansion action.
\method extends this to a state space, whose values are programs, via a stochastic operator $P_\theta$ induced by the LLM:
\begin{equation}
\begin{split}
P_u &\sim P_\theta(\cdot \mid P_v, A_v, s, G), \\
&\text{where } P_\theta = \textsc{LlmEditor}_\theta, \quad s \sim \text{Uniform}(\mathcal{S})
\end{split}
\end{equation}
where $P_\theta$ parameterizes a conditional distribution over modified programs.
$P_\theta$, although parameterized by a pretrained foundation model rather than a model learned using search, functionally serves as a learned component enabling MCTS to operate in complex program spaces.
This aligns with recent work using LLMs in place of learned components for sampling in linear search~\citep{zhou2023large, li2023camel}.
Our \method extends LLM-powered transition kernels from linear to tree search, connecting to learned-transition MCTS and model-based RL~\citep{silver2017mastering, schrittwieser2020mastering}, neural-guided search~\citep{chen2020retro, xu2024reinforcement}, and scientific discovery~\citep{lai2025prim}.

\paragraph{LLM Judge as a Learned Value Function.}
Running $P_u$ yields an execution record $R_u=\textsc{Run}(P_u)$, and the LLM Judge provides a semantic evaluation:
\begin{equation}
(A_u, S_u) = R_\phi(P_u, R_u, G), \text{ } R_\phi = \textsc{LlmJudge}_\phi
\end{equation}
where $R_\phi$ acts as a learned value network.
We denote $S_u = \pi_S(R_\phi(P_u, R_u, G))$, thus $\pi_S$ projects the space of $(A,S)$ onto the space of score $S$.
This aligns with recent works where LLMs serve as noisy reward estimators or evaluators in non-differentiable optimization \citep{zhou2023large, zelikman2022star,shinn2023reflexion}.
Theoretically, $S_u$ constitutes a \emph{bounded, noisy reward}, falling under convergence analysis of MCTS with biased evaluators \citep{lisy2013convergence,efroni2019combine} which shows UCT remains consistent when reward noise is bounded and non-adversarial.
This is the case of our work: scores are bounded and semantic LLM scoring can be treated as non-adversarial.

\paragraph{Theoretical Guarantees Remain an Open Problem.}
In classical MCTS with well-specified MDPs, asymptotic convergence and regret guarantees are established \cite{kocsis2006bandit,bubeck2011xarmed}.
However, for work combining MCTS with LLMs, even though they demonstrate strong empirical performances~\cite{li2025codetree,li2025rethinkmcts,xu2025sra}, they do not focus on theoretical analysis~\cite{katz2024thought}.
For them, the current theoretical work only provides at best partial guarantees (e.g., loose verifier-induced upper bounds \cite{brandfonbrener2024vermcts}), leaving the general problems of \emph{asymptotic convergence} and \emph{regret guarantees} for MCTS and LLM methods as open questions.

}

{

\section{Experiment Details}
\label{appendix:experiment-details}

\subsection{Data Statistics}
\label{appendix:data-statistics}

In Table~\ref{table:data-statistics} we show the statistics of variables in our dataset.   
The maximum number of integrated variables is from the \emph{World Dynamics} system.

\begin{table}[h]
\caption{Statistics of variables in 20 complex systems from our dataset.}
\label{table:data-statistics}
\begin{center}
\begin{small}
\begin{tabular}{lrrr}
\toprule
Variable Type & Mean & Max & Range \\
\midrule
Integrated variables & 29.4 & 69 & 20--69 \\
Intermediate variables (helpers) & 10.9 & 12 & 10--12 \\
\bottomrule
\end{tabular}
\end{small}
\end{center}
\end{table}

\subsection{Expansion Strategies}
\label{appendix:expansion-strategies}

In Table~\ref{table:expansion-strategies}, we show the MCTS expansion strategies.
They are injected into the tree context (see LLM contextualization templates in Appendix~\ref{appendix:llm-contextualization}) for expansion.

\begin{table}[h]
\caption{MCTS expansion strategies.}
\label{table:expansion-strategies}
\begin{tabular}{l p{0.75\linewidth}}
\toprule
Strategy Name & LLM Instructions \\
\midrule
breakthrough & \texttt{Make significant, high-impact changes designed to achieve major performance improvements} \\[0.5em]
aggressive & \texttt{Make bold structural or algorithmic changes} \\[0.5em]
amplify & \texttt{Identify and significantly increase the most promising parameters or mechanisms} \\[0.5em]
exploratory & \texttt{Try completely different parameter combinations} \\[0.5em]
targeted & \texttt{Focus on specific high-impact parameters identified from analysis} \\[0.5em]
contrarian & \texttt{Try approaches opposite to current trends or patterns} \\[0.5em]
balanced & \texttt{Make moderate parameter changes with good risk/reward ratio} \\[0.5em]
conservative & \texttt{Make small, incremental parameter adjustments} \\
\bottomrule
\end{tabular}
\end{table}

\subsection{Scalability and Stability}
\label{appendix:scalability-and-stability}

\paragraph{Scalability}
The design of \method{} allows scaling to large trees (in MCTS) and long execution records.
To do so, in each LLM call, we adaptively subsample the running record of the program of the current node so that the prompt stays within a budget $L$, which we denote as the LLM's maximum context length we would like to impose.
Overall, this preserves the most informative parts of the record while remaining in the overall time complexity of $O(NL)$, where $N$ is the number of expanded nodes for the whole tree search.
This is also reflected in the computational cost and runtime mentioned in Appendix~\ref{appendix:computation-cost-and-runtime}

\paragraph{Stability}
While we provide information regarding the constraints for LLMs to follow, LLMs may not follow such instructions all the time. We found that this is not common but may still happen.
To mitigate it, we have calls to LLM Judge and LLM Editor using structured outputs and at most three retries.
If all attempts fail to respect the constraints, we would discard that expansion and move on to other nodes.
Also, we catch running crashes and numerical issues in records (e.g. NaN) and assign low scores.
This causes the tree search to revert to the parent node and move on with other nodes,
and MCTS naturally backtracks due to the low scores, and such nodes are not revisited frequently.

\subsection{Computational Cost and Runtime}
\label{appendix:computation-cost-and-runtime}

Typically, our experiments have a cost ranging from \$10 to \$50 for each run.
This depends on hyperparameters such as LLM provider, expansion width and search depth,
and reflects a trade-off between exploration depth and resource usage.
Our experiments have a run time of approximately 30 minutes for each run.
The runtime is primarily bounded by LLM calls, rather than local computation.
The cost and runtime are aligned with the design choices for scalability and stability, described in Appendix~\ref{appendix:scalability-and-stability}.
As a result, \method is suitable for real-world applications.

}

{
\section{Experiments: Fitting an Abstract Goal Described in Natural Language}
\label{appendix:experiments-fitting-goal-details}

\begin{figure}[h]
  \figremovespace
  \begin{center}
    \centerline{\includegraphics[width=0.8\columnwidth]{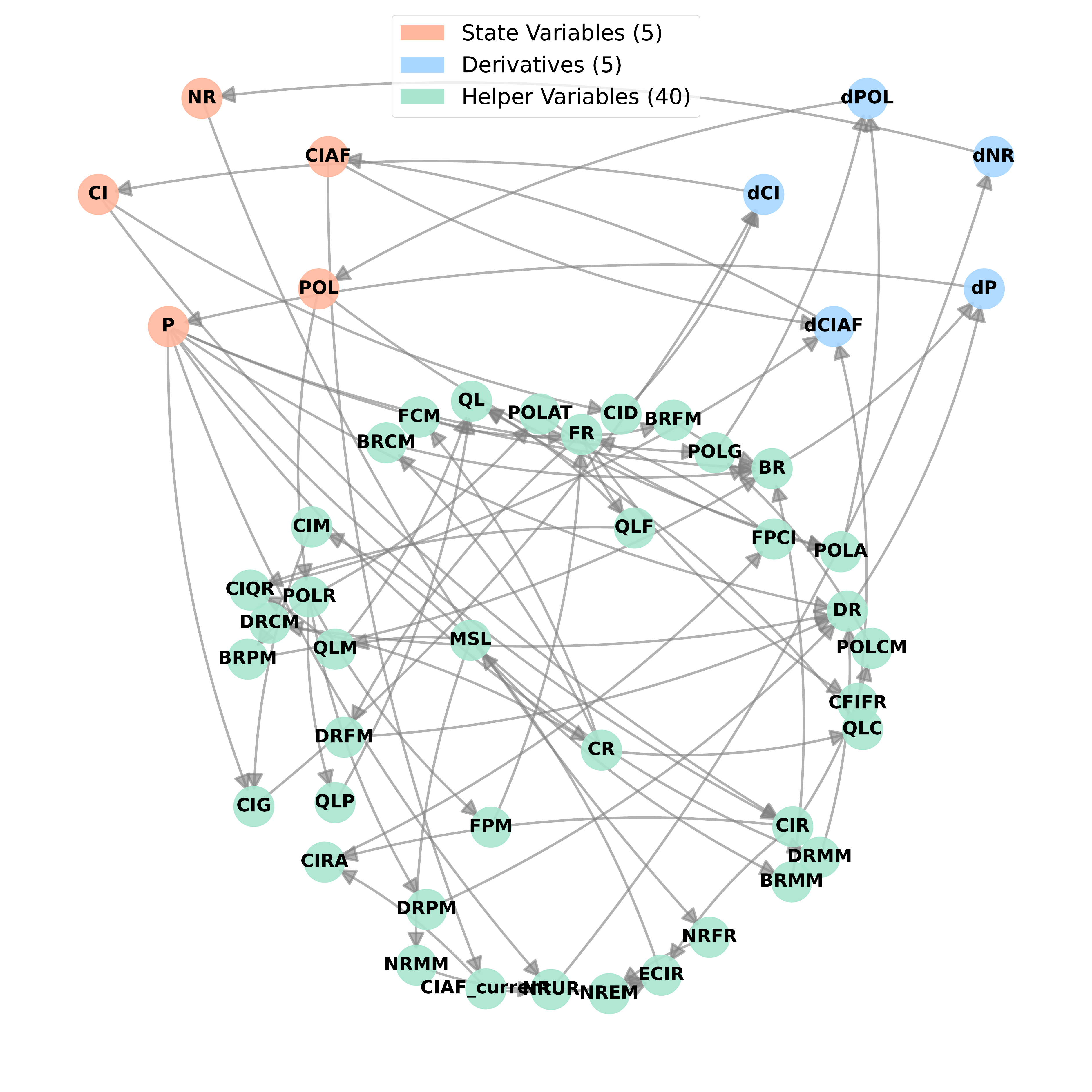}}
    \caption{The feedback structure in World Dynamics System as a dependency graph. The complex and non-linear nature of the dynamics in the complex system is demonstrated by the highly interconnected variables.}
    \label{figure:dependency-graph}
  \end{center}
  \figremovespace
\end{figure}

\begin{table}[h]
  \figremovespace
  \caption{Variable Definitions in World Dynamics System}
  \label{table:dependency-graph-definitions}
  \begin{center}
    \resizebox{0.5\textwidth}{!}{%
      \begin{tabular}{llll}
        \hline
        \textbf{Abbr.} & \textbf{Meaning} & \textbf{Abbr.} & \textbf{Meaning} \\
        \hline
        BR & Birth Rate & FPCI & Food Potential From Capital Investment \\
        BRCM & Birth Rate From Crowding Multiplier & FPM & Food From Pollution Multiplier \\
        BRFM & Birth Rate From Food Multiplier & FR & Food Ratio \\
        BRMM & Birth Rate From Material Multiplier & MSL & Material Standard Of Living \\
        BRPM & Birth Rate From Pollution Multiplier & NR & Natural Resources \\
        CFIFR & Capital Fraction Indicated By Food Ratio & NREM & Natural Resource Extraction Multiplier \\
        CI & Capital Investment & NRFR & Natural Resource Fraction Remaining \\
        CIAF & Capital Investment In Agriculture Fraction & NRMM & Natural Resource From Material Multiplier \\
        CID & Capital Investment Discard & NRUR & Natural Resource Usage Rate \\
        CIG & Capital Investment Generation & P & Population \\
        CIM & Capital Investment Multiplier & POL & Pollution \\
        CIQR & Capital Investment From Quality Ratio & POLA & Pollution Absorption \\
        CIR & Capital Investment Ratio & POLAT & Pollution Absorption Time \\
        CIRA & Capital Investment Ratio In Agriculture & POLCM & Pollution From Capital Multiplier \\
        CR & Crowding Ratio & POLG & Pollution Generation \\
        DR & Death Rate & POLR & Pollution Ratio \\
        DRCM & Death Rate From Crowding Multiplier & QL & Quality Of Life \\
        DRFM & Death Rate From Food Multiplier & QLC & Quality Of Life From Crowding \\
        DRMM & Death Rate From Material Multiplier & QLF & Quality Of Life From Food \\
        DRPM & Death Rate From Pollution Multiplier & QLM & Quality Of Life From Material \\
        ECIR & Effective Capital Investment Ratio & QLP & Quality Of Life From Pollution \\
        FCM & Food From Crowding Multiplier & dCI & $\Delta$ Capital Investment \\
        dP & $\Delta$ Population & dCIAF & $\Delta$ Capital Investment In Agriculture Fraction \\
        dPOL & $\Delta$ Pollution & dNR & $\Delta$ Natural Resources \\
        \hline
      \end{tabular}%
    }
  \end{center}
  \figremovespace
\end{table}

\begin{figure}[h!]
  \centering
  \includegraphics[width=0.8\columnwidth,trim=0 160pt 400pt 0,clip]{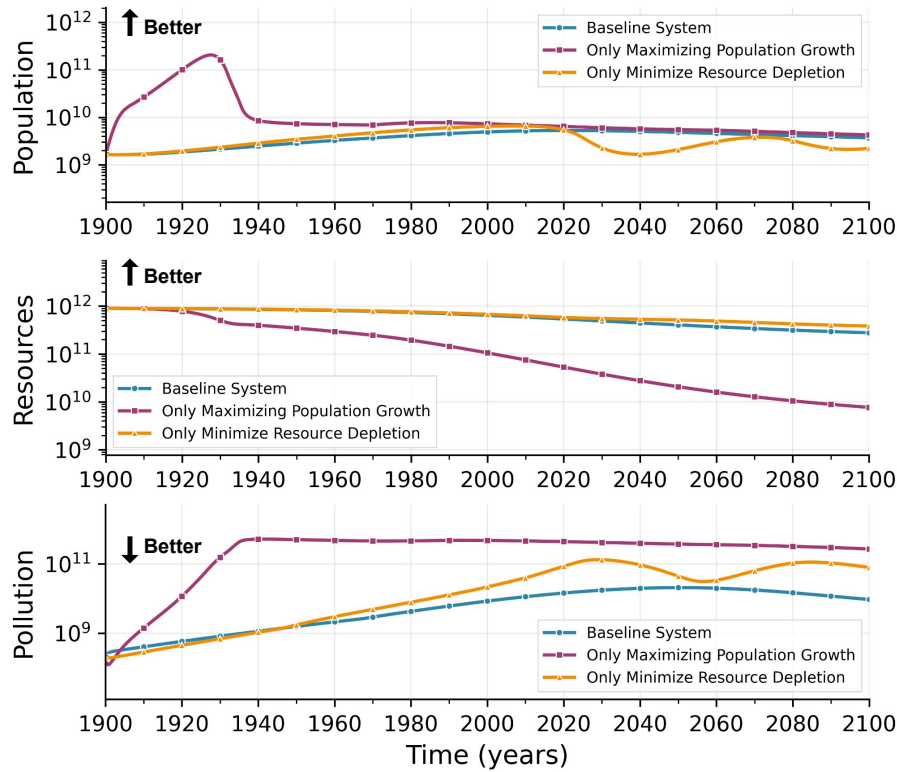}
  \caption{Running record from optimization targeting individual goals. Focusing on a single objective often is at the expense of other goals, showing the intrinsic challenges in system optimization with multiple subgoals.}
  \label{figure:fitting-single-goal}
\end{figure}

\subsection{World Dynamics Complex System Details}
\label{appendix:world-0dynamics-complex-system-details}

In this experiment, we study a complex system that models the co-evolution of human population, resource utilization and pollution.
As with many complex systems, it is a non-linear and feedback-driven system.
The baseline is the \emph{published system} in the monograph~\cite{forrester1971world}.
As shown in Figure~\ref{figure:dependency-graph} (definitions in Table~\ref{table:dependency-graph-definitions}), this system demonstrates a significant structural complexity and non-linearity.

\subsection{Natural Language Goal for Optimizing Complex Systems}
\label{appendix:natural-language-goal-for-optimizing-complex-systems}

Our goal for this experiment is to jointly balance the population growth, resource usage, and pollution. The exact natural-language goal used in the experiment in Section~\ref{section:experiments-fitting-goal} is provided below.

\begin{tcolorbox}[colback=blue!5!white,colframe=blue!50!white,title={Goal for balancing population growth, resource usage, and pollution},breakable]
\begin{lstlisting}[style=goal]
Balance population, resources, and environment by optimizing to (1) maximize population growth, (2) minimize the resource depletion rate, and (3) minimize the pollution accumulation rate. Seek the best trade-off where the population grows sustainably without depleting resources too quickly or creating excessive pollution. (Important: only change the coefficients in the helper. Do not change any coefficient by more than $50\\%$ to prevent variable explosion.)
\end{lstlisting}
\end{tcolorbox}

\subsection{Intrinsic Challenges of Optimizing Complex Systems with Competing Subgoals}
\label{appendix:intrinsic-challenges-of-optimizing-complex-systems-with-competing-subgoals}

Optimizing this system is highly non-trivial for two reasons:
(1) with such interdependencies among variables, a single change can trigger cascading effects multiple time steps away, and
(2) Often, the goal specified in natural language could be vague, and the decomposition of it into subgoals requires bridging the gap between goal and understanding the interactions among system variables. As a result, subgoals can be competing: for example, increasing population typically requires greater resource consumption.
(3) The baseline as a \emph{resulting, published system} is strong enough to reach a kind of Pareto frontier, such that any further optimizing of one subgoal often comes at the expense of other subgoals.
Overall, this means that optimization in this experiment is substantially challenging, which we will detail in the following text.

To demonstrate the intrinsic challenges brought by competing subgoals,
we conduct experiments where we ask our method to optimize for each single subgoal.
Concretely, we consider two cases: (1) optimizing to only ``maximize population growth'' and (2) optimizing to only ``minimize resource depletion''.
We show in Figure~\ref{figure:fitting-single-goal} three key variables:  population and resources (in goals) and pollution (an additional variable) for these two cases along with the baseline system.
Compared to the baseline, case (1) increases population growth but has worse resource availability and pollution levels. Similarly, case (2) decreases resource depletion but has worse population and pollution levels.
Such observations show that the intricate interdependencies (see the feedback structures described above) lead to challenges that arise not only from the conflicting objectives but also from the intrinsic properties of the system that mandate the trade-offs.
It also highlights the challenges of the experiments and the capabilities of our method \method{}.

}

{

\section{Experiment: Fitting a Concrete Record}

\label{appendix:experiments-fitting-record-details}

\subsection{Ground Truth System}
\label{appendix:experiments-fitting-record-details-ground-truth-system}

We provide the implementation details for the population growth model used as the ground truth system in this experiment (Section~\ref{section:experiments-fitting-record}).
This complex system models the population dynamics with stochastic components, using random fluctuations,
shown below:

\begin{lstlisting}[style=code]
# ===================== BEGIN CONFIG =========================
# Simulation configuration parameters (DO NOT CHANGE - these are set by the system)
dt = 0.1  # time step size
t0 = 0.0  # start time
tf = 3500.0  # end time
# ====================== END CONFIG ==========================

# ====================== BEGIN STATE =========================
# STATE VARIABLES - integrated over time (must define derivatives below)
POPULATION = 2.0  # NECESSARY
SUM_POP = 0.0  # NECESSARY
# ======================= END STATE ==========================

# ======================= MAIN LOOP ==========================
t = t0
while t <= tf + 1e-12:

    # =================== BEGIN HELPERS ======================
    # HELPERS - All computed variables (algebraic and intermediate expressions)
    # in dependency order. These can also use wrappers like graph(), sin(), exp(),
    # delay(), smth1() - e.g. delayed/smoothed signals.

    # Birth rate calculation
    BIRTHS = 0.07 * POPULATION

    # Nominal death rate calculation
    NOMINAL_DR = (exp(-0.01 * t) * 0.03 + 0.01) * 1 + 0.04 * 0

    # Death rate distribution with normal random variation
    DR_DISTRIBUTION = normal(NOMINAL_DR, 0.005 * POPULATION)

    # Control death rate within bounds
    DR_DIST_CONTROL = DR_DISTRIBUTION if (DR_DISTRIBUTION >= 0.01 and DR_DISTRIBUTION <= 1) else 0.01

    # Final death rate calculation
    DEATH_RATE = (
        (DR_DIST_CONTROL if DR_DIST_CONTROL > NOMINAL_DR else NOMINAL_DR) * 1 + 0 * DR_DIST_CONTROL + 0 * NOMINAL_DR
    )

    # Deaths calculation
    DEATHS = DEATH_RATE * POPULATION

    # Current population flow (conditional on time)
    CURRENT_POP = POPULATION if t > 100 else 0

    # Average population calculation
    AVG_POP = SUM_POP / (t - 100) if t != 100 else 0
    # ==================== END HELPERS =======================

    # ================= BEGIN DERIVATIVES ====================
    dPOPULATION = BIRTHS - DEATHS
    dSUM_POP = CURRENT_POP
    # ================== END DERIVATIVES =====================

    # --------------- Euler integration (engine; do not edit) ---------------
    POPULATION = POPULATION + dt * dPOPULATION
    SUM_POP = SUM_POP + dt * dSUM_POP
    t += dt
\end{lstlisting}

\subsection{Baseline Systems}
\label{appendix:experiments-fitting-record-details-baseline-systems}

In evaluation, we compare our \method against black-box optimization using Optuna.
We provide three levels of formula support for it, each increasing in complexity and thus providing advantages.

\paragraph{Level 1: No Formulae Baseline}

This no formula baseline provides only two simple constant parameters, namely \texttt{BIRTH\_RATE} and \texttt{DEATH\_RATE}.
This represents the most challenging scenario for the baseline, since the optimization method, in capturing the dynamics of the ground truth system, can rely only on these parameters without any sophisticated formula structures and thus needs to come up with the formulae on its own.

\begin{lstlisting}[style=code]
BIRTH_RATE = 0.001
DEATH_RATE = 0.001

...
while t <= tf + 1e-12:
    # Birth rate calculation
    BIRTHS = BIRTH_RATE * POPULATION

    # Deaths calculation
    DEATHS = DEATH_RATE * POPULATION

    ...
\end{lstlisting}

\paragraph{Level 2: Simple Formulae Baseline}
This simple formulae baseline provides a basic feedback structure in the form of population-dependent death rate.
This structure helps capture some of the system's behavior that is self-regulating.

\begin{lstlisting}[style=code]
BIRTH_RATE = 0.001
DEATH_RATE = 0.001
EFFECTIVE_DEATH_RATE_COEFF = 0.001
MAX_POPULATION = 300

...
while t <= tf + 1e-12:
    # Birth rate calculation
    BIRTHS = BIRTH_RATE * POPULATION

    # Deaths calculation
    EFFECTIVE_DEATH_RATE = max(DEATH_RATE, EFFECTIVE_DEATH_RATE_COEFF * max(1.0, POPULATION / MAX_POPULATION))
    DEATHS = EFFECTIVE_DEATH_RATE * POPULATION

    ...
\end{lstlisting}

\paragraph{Level 3: Full Formulae Baseline}
This full formulae baseline provides the complete formulae structure that is the same as the ground truth system, thus providing its sophisticated dynamics.
For this baseline, the optimization method only needs to tune the coefficient parameters, without needing to discover the formulae feedbacks.
This provides the most advantage for the baseline method, since it essentially simplifies the task to parameter fitting.

\begin{lstlisting}[style=code]
BIRTH_RATE = 0.001
NOMINAL_DR_EXP_COEFF_T = -0.001 # should be negative to avoid explosion.
NOMINAL_DR_EXP_SCALE = 0.001
NOMINAL_DR_EXP_SHIFT = 0.001
DR_DISTRIBUTION_STD_COEFF_POPULATION = 0.001

...
while t <= tf + 1e-12:
    # Birth rate calculation
    BIRTHS = BIRTH_RATE * POPULATION

    # Nominal death rate calculation
    NOMINAL_DR = (exp(NOMINAL_DR_EXP_COEFF_T * t) * NOMINAL_DR_EXP_SCALE + NOMINAL_DR_EXP_SHIFT)

    # Death rate distribution with normal random variation
    DR_DISTRIBUTION = normal(NOMINAL_DR, DR_DISTRIBUTION_STD_COEFF_POPULATION * POPULATION)

    # Control death rate within bounds
    DR_DIST_CONTROL = DR_DISTRIBUTION if (DR_DISTRIBUTION >= 0.01 and DR_DISTRIBUTION <= 1) else 0.01

    # Final death rate calculation
    DEATH_RATE = (
        (DR_DIST_CONTROL if DR_DIST_CONTROL > NOMINAL_DR else NOMINAL_DR) * 1
    )

    # Deaths calculation
    DEATHS = DEATH_RATE * POPULATION

    ...
\end{lstlisting}

As we show in Section~\ref{section:experiments-fitting-record},
\method remains better in terms of performance.
Even without scaffolding of formulae structure,  \method discovers structure and parameters and thus models the dynamics from the record.

\subsection{Dynamic Time Warping Distance}
\label{dynamic-time-warping-distance}

For metrics, we use Dynamic Time Warping (DTW)~\cite{sakoe1978dynamic} in addition to L1 distance.
While L1 distance measures point-wise accuracy, DTW measures trajectory similarity under optimal, temporal alignment.
In DTW, we use \texttt{dtaidistance} library~\cite{meert2020wannesm} and apply a Sakoe-Chiba band constraint with window size $w=250$,
which restricts the warping path to stay within 205 time steps.
This window size corresponds to $7.1\%$ of our sequence length ($3500$ time steps) and falls within the recommended optimal range of $5-10\%$ that prevents issues in warping alignments, while preserving the quality of alignments~\cite{ratanamahatana2004everything}.

\subsection{Stochasticity in Ground Truth Complex System}
\label{appendix:stochasticity-in-ground-truth-complex-system}

We note that the peaks and dips observed in the ground truth system record (trajectories) 
arise from the stochasticity in the death-rate process: 
In the ground truth system, the death rate is sampled at every time step from a normal distribution parameterized by the current population (see Line 16 of Level 3 code above in Appendix~\ref{appendix:experiments-fitting-record-details-ground-truth-system}), which introduces stochasticity into the complex system:

\texttt{DR\_DISTRIBUTION = normal(NOMINAL\_DR, 0.005 * POPULATION)} 

To quantify the magnitude of such stochasticity, we run the ground truth system with $10$ different random seeds and study the variance.
As Figure~\ref{figure:gt-seeds-statistics} shows, the record (trajectory) exhibits variation ranges from this stochasticity;
thus, the volatility is an intrinsic property of the system rather than an artifact of the optimizer.
Once we account for this stochasticity, we can conclude that \method{}'s record generally falls within the ground truth variability ranges, and it can capture the magnitude and time of the major peaks and dips.

\begin{figure}[h!]
\begin{center}
\includegraphics[width=0.425\textwidth]{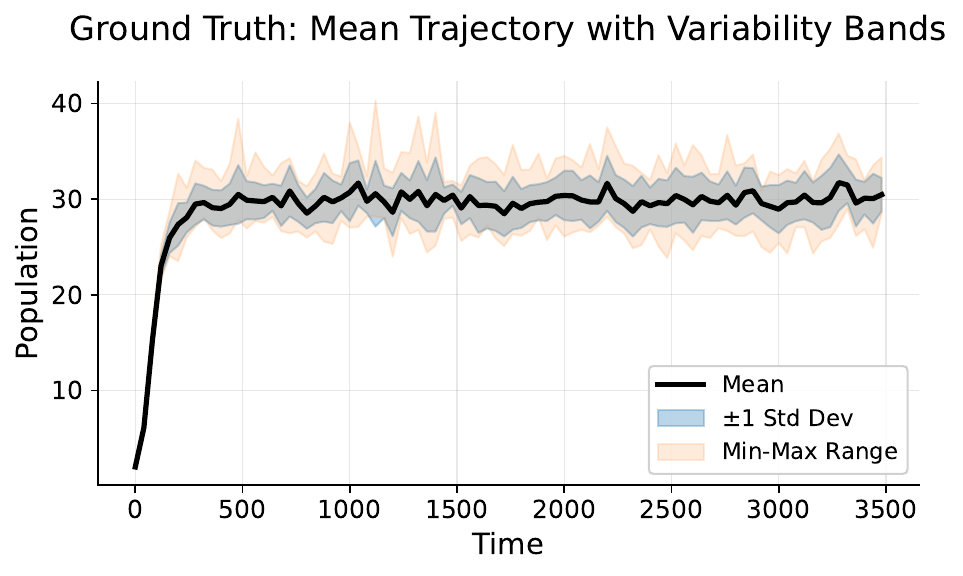}
\includegraphics[width=0.425\textwidth]{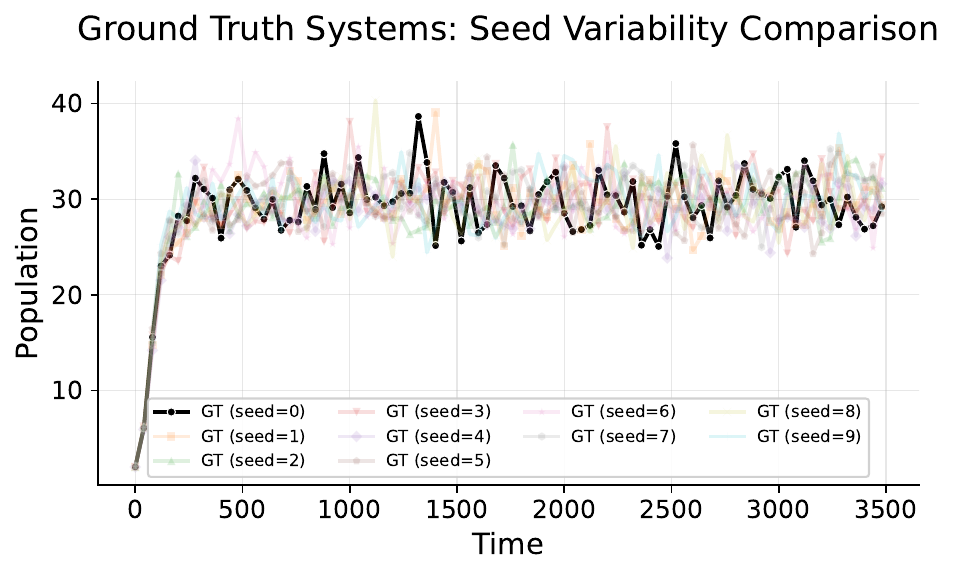}
\caption{Visualization of ground truth system record (trajectories) across 10 different random seeds, showing the volatility induced by stochastic death-rate sampling (left) and comparison between these 10 runs (right).}
\label{figure:gt-seeds-statistics}
\end{center}
\end{figure}

\subsection{Optuna's Performance on Baseline with Full Formulae}
\label{appendix:optunas-performance-on-baseline-with-full-formulae}

We use $100$ trials for Optuna in the experiments in contrast to the case of \method{} where we use $10$ MCTS iterations.
We do so primarily because each MCTS iteration is significantly more expensive, both in terms of computation cost and runtime, than a single Optuna trial.
This gives the baselines a stronger opportunity to match the dynamics.

We particularly look into the baseline where Optuna has full formulae support, which is the least challenging for it as it only needs to fit parameters.
In theory, with full formulae access and a sufficient number of trials, Optuna could approach the ground truth system's parameters.
In practice, however, we observe that it does not perfectly match the dynamics of the ground truth system:
As shown in Table~\ref{table:performance-comparison-bl-easy} and Figure~\ref{figure:fitting-comparison-bl-easy},
we add an extra run of Optuna (``Run 2'', in addition to ``Run 1'' which is reported in Section~\ref{section:experiments-fitting-record}), which does not lead to better performance.

Considering the stochasticity above, we believe it could be explained by two factors:
(1) the ``noise floor'' in the ground truth system, which can be visually verified by observing that peaks and dips are not in sync across runs, and (2) the optimization landscape is challenging for purely numeric search, where the parameter space is high-dimensional and sensitive such that small changes can lead to qualitatively different records due to non-linear feedbacks.

\begin{minipage}[t]{0.55\textwidth}
    \centering
    \small
    \captionof{table}{L1 Distance and DTW comparison.}

    \definecolor{color1}{HTML}{006666}
    \definecolor{color2}{HTML}{AA3366}
    \definecolor{color3}{HTML}{5533AA}
    \definecolor{color4}{HTML}{CC4400}
    \definecolor{color5}{HTML}{00AA44}
    \begin{tabular}{p{1.6cm}p{0.12cm}p{0.12cm}p{0.12cm}rr}
        \toprule
        \multirow{2}{1.6cm}{\centering Method} & \multicolumn{3}{c}{Formulae} & \multirow{2}{*}{\centering L1} & \multirow{2}{*}{\centering DTW} \\
        & No & S. & F. & & \\
        \cmidrule{2-4}
        \midrule
        \textcolor{color3}{Optuna \textbf{(Run 1)}} & & & \textcolor{color3}{$\checkmark$} & \textcolor{color3}{$3.71$} & \textcolor{color3}{${477.52}$} \\
        \textcolor{color3}{Optuna \textbf{(Run 2)}} & & & \textcolor{color3}{$\checkmark$} & \textcolor{color3}{$4.26$} & \textcolor{color3}{${605.05}$} \\
        \textcolor{color5}{\method (GPT-5.1, Ours)} & \textcolor{color5}{$\checkmark$} & & & \textcolor{color5}{$\mathbf{2.22}$} & \textcolor{color5}{$\mathbf{433.13}$} \\
        \bottomrule
    \end{tabular}
    \label{table:performance-comparison-bl-easy}
    \figremovespace
\end{minipage}
\hfill
\begin{minipage}[t]{0.45\textwidth}
    \centering
    \raisebox{\dimexpr-\height+\ht\strutbox\relax}{\includegraphics[width=\linewidth]{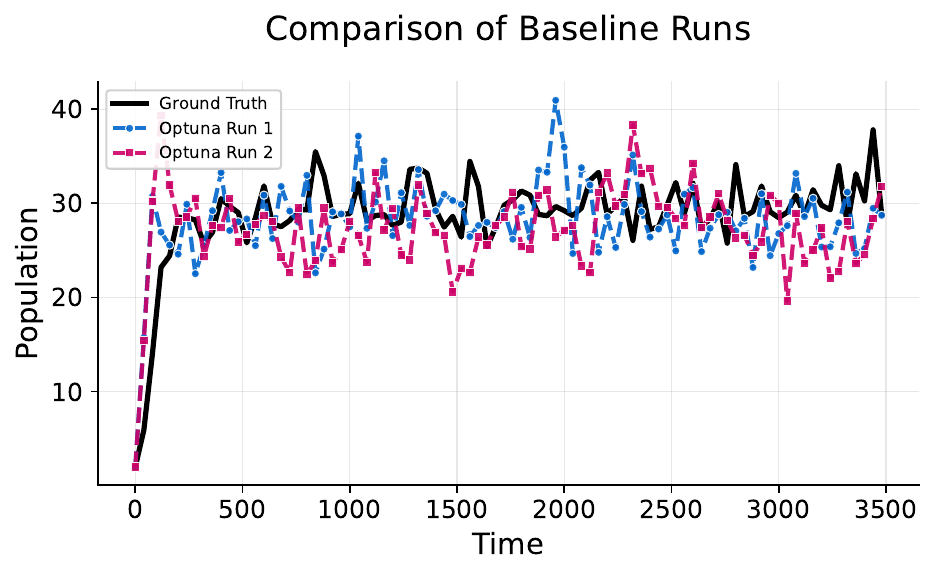}}
    \captionof{figure}{Comparison of Baseline Runs.}
    \label{figure:fitting-comparison-bl-easy}
    \figremovespace
\end{minipage}

}

{

\section{Interpretability Analysis}

\label{appendix:interpretability-details}

\subsection{Relation between LLM reasoning and LLM scores}

We observe, with human inspection, that the LLM Judge score and the reasoning are aligned and there is no systematic divergence. Notably, the reasoning helps produce better edits while the score helps MCTS node selection, and their roles are complementary.

\subsection{LLM Analysis Examples}

We show the complete responses of LLM Judge and LLM Editor from an actual run in Section~\ref{section:experiments-fitting-goal} below.
The LLM Judge evaluates the system against the given goal and explains the observed strengths, weaknesses, and trade-offs.
The LLM Editor in turn proposes new code to better achieve the goal.
This demonstrates the interpretability and reasoning capabilities of \method{}.
We manually colored text related to \textcolor{populationcolor}{Population}, \textcolor{pollutioncolor}{pollution}, and \textcolor{resourcecolor}{resources} for visibility.
Note that there is no Editor Response 1, which is the initial system given as is.

}

\begin{tcolorbox}[colback=blue!5!white,colframe=blue!50!white,title={Judge Response 1},breakable,pad at break*=1mm,enlarge top at break by=0pt,enlarge bottom at break by=0pt]
\begin{lstlisting}[
    language={},basicstyle=\small\ttfamily,breaklines=true,breakindent=0pt,backgroundcolor={},numbers=none,frame=none,escapeinside={(*@}{@*)},
    literate={population}{{\textcolor{populationcolor}{population}}}{10}
             {Population}{{\textcolor{populationcolor}{Population}}}{10}
             {POPULATION}{{\textcolor{populationcolor}{POPULATION}}}{10}
             {resources}{{\textcolor{resourcecolor}{resources}}}{9}
             {Resources}{{\textcolor{resourcecolor}{Resources}}}{9}
             {resource}{{\textcolor{resourcecolor}{resource}}}{8}
             {Resource}{{\textcolor{resourcecolor}{Resource}}}{8}
             {RESOURCES}{{\textcolor{resourcecolor}{RESOURCES}}}{9}
             {RESOURCE}{{\textcolor{resourcecolor}{RESOURCE}}}{8}
             {pollution}{{\textcolor{pollutioncolor}{pollution}}}{9}
             {Pollution}{{\textcolor{pollutioncolor}{Pollution}}}{9}
             {POLLUTION}{{\textcolor{pollutioncolor}{POLLUTION}}}{9}
]
The simulation shows a world system model tracking (*@\textcolor{populationcolor}{population}@*), (*@\textcolor{resourcecolor}{resources}@*), capital investment, and (*@\textcolor{pollutioncolor}{pollution}@*) from 1900-2100. Key findings: (*@\textcolor{populationcolor}{Population grows from 1.65B to 3.7B}@*), (*@\textcolor{resourcecolor}{natural resources decline from 900B to 278B units (69\% depletion)}@*), capital investment increases from 400M to 6B units, and (*@\textcolor{pollutioncolor}{pollution rises from 200M to 9.3B units (46x increase)}@*).

Analyzing against the three optimization objectives:

1. MAXIMIZE (*@\textcolor{populationcolor}{POPULATION GROWTH}@*): (*@\textcolor{populationcolor}{Population grows 124\% over 200 years (1.65B to 3.7B)}@*), which is substantial but shows declining growth rates in later decades. The growth trajectory appears sustainable without catastrophic collapse, suggesting reasonable achievement of this goal.

2. MINIMIZE (*@\textcolor{resourcecolor}{RESOURCE DEPLETION RATE}@*): (*@\textcolor{resourcecolor}{Natural resources decline from 900B to 278B units, representing a 69\% depletion over 200 years}@*). The depletion rate averages about 3.1B units per year, which is quite high. The (*@\textcolor{resourcecolor}{natural resource fraction remaining (NRFR) drops to 0.309 by 2100, indicating severe resource stress}@*). This objective is poorly achieved.

3. MINIMIZE (*@\textcolor{pollutioncolor}{POLLUTION ACCUMULATION RATE}@*): (*@\textcolor{pollutioncolor}{Pollution increases dramatically from 200M to 9.3B units, a 46-fold increase}@*). The (*@\textcolor{pollutioncolor}{pollution ratio (POLR) reaches 2.587 by 2100, indicating severe environmental degradation}@*). This represents a complete failure to minimize (*@\textcolor{pollutioncolor}{pollution accumulation}@*).

The model shows classic overshoot behavior - initial growth followed by (*@\textcolor{resourcecolor}{resource constraints}@*) and (*@\textcolor{pollutioncolor}{environmental degradation}@*). While (*@\textcolor{populationcolor}{population doesn't collapse entirely}@*), the system exhibits unsustainable patterns with massive (*@\textcolor{resourcecolor}{resource depletion}@*) and (*@\textcolor{pollutioncolor}{pollution accumulation}@*). The trade-offs are heavily skewed toward short-term (*@\textcolor{populationcolor}{population growth}@*) at the expense of long-term sustainability.

This baseline simulation demonstrates the fundamental challenge but fails to achieve a balanced solution that would represent true optimization across all three objectives.
\end{lstlisting}
\end{tcolorbox}

\begin{tcolorbox}[colback=green!5!white,colframe=blue!50!white,title={Editor Response 2},breakable]
\begin{lstlisting}[
    language={},basicstyle=\small\ttfamily,breaklines=true,breakindent=0pt,backgroundcolor={},numbers=none,frame=none,escapeinside={(*@}{@*)},
    literate={population}{{\textcolor{populationcolor}{population}}}{10}
             {Population}{{\textcolor{populationcolor}{Population}}}{10}
             {POPULATION}{{\textcolor{populationcolor}{POPULATION}}}{10}
             {resources}{{\textcolor{resourcecolor}{resources}}}{9}
             {Resources}{{\textcolor{resourcecolor}{Resources}}}{9}
             {resource}{{\textcolor{resourcecolor}{resource}}}{8}
             {Resource}{{\textcolor{resourcecolor}{Resource}}}{8}
             {RESOURCES}{{\textcolor{resourcecolor}{RESOURCES}}}{9}
             {RESOURCE}{{\textcolor{resourcecolor}{RESOURCE}}}{8}
             {pollution}{{\textcolor{pollutioncolor}{pollution}}}{9}
             {Pollution}{{\textcolor{pollutioncolor}{Pollution}}}{9}
             {POLLUTION}{{\textcolor{pollutioncolor}{POLLUTION}}}{9}
]
BREAKTHROUGH APPROACH: Implementing aggressive (*@\textcolor{resourcecolor}{resource conservation}@*) and (*@\textcolor{pollutioncolor}{pollution control}@*) measures to achieve sustainable (*@\textcolor{populationcolor}{population growth}@*). The current simulation shows unsustainable trends - (*@\textcolor{populationcolor}{population grows to 3.7B by 2100}@*) while (*@\textcolor{resourcecolor}{natural resources decline from 900B to 278B units (69\% depletion)}@*) and (*@\textcolor{pollutioncolor}{pollution increases dramatically from 200M to 13.5B units (67x increase)}@*). This represents poor balance between the three objectives.

My strategy focuses on three key interventions:
1. (*@\textcolor{resourcecolor}{RESOURCE EFFICIENCY: Reducing natural resource usage multiplier coefficients by 25-40\% to slow depletion}@*)
2. (*@\textcolor{pollutioncolor}{POLLUTION CONTROL: Reducing pollution generation coefficients by 30-40\% and improving absorption rates by 25\%}@*)  
3. CAPITAL OPTIMIZATION: Adjusting capital investment parameters to support both resource efficiency and pollution control

Specific modifications:
- Reduce base natural resource usage rate from 1.0 to 0.7 (30% improvement)
- Reduce base pollution generation rate from 1.0 to 0.6 (40% reduction)
- Reduce capital investment discard rate from 0.025 to 0.02 (20% improvement in capital retention)
- Reduce base death rate from 0.028 to 0.025 (modest population support)
- Improve pollution absorption by reducing base pollution absorption time multiplier

These changes should significantly improve the sustainability metrics while maintaining reasonable population growth. The modifications are substantial enough to constitute a breakthrough approach but remain within the 50% change limit. Expected outcomes: (*@\textcolor{resourcecolor}{slower resource depletion (targeting <50\% by 2100)}@*), (*@\textcolor{pollutioncolor}{dramatically reduced pollution accumulation (targeting <5B units)}@*), and (*@\textcolor{populationcolor}{sustained population growth to 4-5B people}@*).

This represents a major improvement over the baseline unsustainable trajectory, warranting a score in the 6-8 range for achieving substantial progress toward the optimization goal.
\end{lstlisting}
\end{tcolorbox}

\begin{tcolorbox}[colback=blue!5!white,colframe=blue!50!white,title={Judge Response 2},breakable]
\begin{lstlisting}[
    language={},basicstyle=\small\ttfamily,breaklines=true,breakindent=0pt,backgroundcolor={},numbers=none,frame=none,escapeinside={(*@}{@*)},
    literate={population}{{\textcolor{populationcolor}{population}}}{10}
             {Population}{{\textcolor{populationcolor}{Population}}}{10}
             {POPULATION}{{\textcolor{populationcolor}{POPULATION}}}{10}
             {resources}{{\textcolor{resourcecolor}{resources}}}{9}
             {Resources}{{\textcolor{resourcecolor}{Resources}}}{9}
             {resource}{{\textcolor{resourcecolor}{resource}}}{8}
             {Resource}{{\textcolor{resourcecolor}{Resource}}}{8}
             {RESOURCES}{{\textcolor{resourcecolor}{RESOURCES}}}{9}
             {RESOURCE}{{\textcolor{resourcecolor}{RESOURCE}}}{8}
             {pollution}{{\textcolor{pollutioncolor}{pollution}}}{9}
             {Pollution}{{\textcolor{pollutioncolor}{Pollution}}}{9}
             {POLLUTION}{{\textcolor{pollutioncolor}{POLLUTION}}}{9}
]
This code variant shows significant improvements in achieving sustainable (*@\textcolor{populationcolor}{population growth}@*) while reducing (*@\textcolor{resourcecolor}{resource depletion}@*) and (*@\textcolor{pollutioncolor}{pollution}@*). The simulation results demonstrate several key achievements:

1. **(*@\textcolor{populationcolor}{Population Growth}@*)**: (*@\textcolor{populationcolor}{Population grows from 1.65B to 5.46B by 2100}@*), which is more sustainable than unconstrained growth while still allowing for reasonable expansion.

2. **(*@\textcolor{resourcecolor}{Resource Conservation}@*)**: (*@\textcolor{resourcecolor}{Natural resources decline from 900B to 262.7B units, representing a 70\% depletion rather than complete exhaustion}@*). This is a substantial improvement over the parent node which showed more severe (*@\textcolor{resourcecolor}{resource depletion}@*).

3. **(*@\textcolor{pollutioncolor}{Pollution Control}@*)**: (*@\textcolor{pollutioncolor}{Pollution levels peak around 2070 at ~14.7B units then stabilize}@*), showing the system can manage (*@\textcolor{pollutioncolor}{pollution accumulation}@*) rather than allowing runaway growth.

4. **System Stability**: The key indicators show the system reaching a more stable state by 2100:
   - Birth rates and death rates converge to sustainable levels
   - Resource usage rates stabilize
   - Capital investment maintains positive growth
   - Quality of life metrics remain reasonable

5. **Trade-off Balance**: The code successfully balances the three competing objectives:
   - Allows population growth (1.65B -> 5.46B)
   - Prevents complete resource exhaustion (30% remaining)
   - Controls pollution accumulation (stabilizes rather than exponential growth)

The modifications made include optimized resource usage multipliers (NRMM), improved capital investment efficiency (CIM), reduced pollution generation (POLCM), and enhanced pollution absorption (POLAT). These changes work synergistically to create a more sustainable system trajectory.

Compared to the parent node (score 3.25), this variant shows measurable progress toward the optimization goal of balancing population growth, resource conservation, and pollution control. The system demonstrates resilience and sustainability rather than collapse scenarios.
\end{lstlisting}
\end{tcolorbox}

\begin{tcolorbox}[colback=green!5!white,colframe=blue!50!white,title={Editor Response 3},breakable]
\begin{lstlisting}[
    language={},basicstyle=\small\ttfamily,breaklines=true,breakindent=0pt,backgroundcolor={},numbers=none,frame=none,escapeinside={(*@}{@*)},
    literate={population}{{\textcolor{populationcolor}{population}}}{10}
             {Population}{{\textcolor{populationcolor}{Population}}}{10}
             {POPULATION}{{\textcolor{populationcolor}{POPULATION}}}{10}
             {resources}{{\textcolor{resourcecolor}{resources}}}{9}
             {Resources}{{\textcolor{resourcecolor}{Resources}}}{9}
             {resource}{{\textcolor{resourcecolor}{resource}}}{8}
             {Resource}{{\textcolor{resourcecolor}{Resource}}}{8}
             {RESOURCES}{{\textcolor{resourcecolor}{RESOURCES}}}{9}
             {RESOURCE}{{\textcolor{resourcecolor}{RESOURCE}}}{8}
             {pollution}{{\textcolor{pollutioncolor}{pollution}}}{9}
             {Pollution}{{\textcolor{pollutioncolor}{Pollution}}}{9}
             {POLLUTION}{{\textcolor{pollutioncolor}{POLLUTION}}}{9}
]
EXPLORATORY APPROACH: Implementing radical efficiency improvements through enhanced capital productivity, aggressive pollution reduction, and optimized resource conservation to achieve breakthrough sustainability.

Looking at the tree context, I'm at node 0_0 with a score of 6.75, which is tied for the global best. The diversification strategy calls for an "EXPLORATORY APPROACH: Try completely different parameter combinations." The current simulation shows (*@\textcolor{populationcolor}{population growing from 1.65B to 5.46B (3.3x increase)}@*) while (*@\textcolor{resourcecolor}{resources decline from 900B to 262B (71\% depletion)}@*) and (*@\textcolor{pollutioncolor}{pollution increases dramatically from 200M to 14.3B (71x increase)}@*).

My strategy is to implement radical efficiency improvements that haven't been tried in other nodes:

1. **Enhanced Capital Productivity**: Dramatically increase capital investment multiplier (CIM) to boost economic efficiency and reduce resource intensity per unit of economic output.

2. **Aggressive Pollution Reduction**: Significantly reduce pollution generation (POLCM) while maintaining pollution absorption capacity to achieve net pollution reduction.

3. **Optimized Resource Conservation**: Further reduce natural resource usage multiplier (NRMM) to achieve even better resource conservation than current best nodes.

4. **Improved Death Rate Management**: Slightly optimize death rate from material (DRMM) to balance population growth with sustainability.

5. **Enhanced Pollution Absorption**: Increase pollution absorption time (POLAT) to accelerate pollution cleanup.

This combination should achieve:
- More sustainable (*@\textcolor{populationcolor}{population growth (target ~4-5B by 2100)}@*)
- Better (*@\textcolor{resourcecolor}{resource conservation (target >40\% remaining)}@*)
- Dramatic (*@\textcolor{pollutioncolor}{pollution reduction (target <5B by 2100)}@*)

Expected improvements over current 6.75 score:
- (*@\textcolor{populationcolor}{Population growth}@*): Moderate improvement through better balance
- (*@\textcolor{resourcecolor}{Resource depletion}@*): Significant improvement through enhanced efficiency
- (*@\textcolor{pollutioncolor}{Pollution control}@*): Major breakthrough through aggressive reduction

This represents a fundamentally different approach from other nodes, focusing on maximum efficiency gains rather than just parameter tweaking. I expect this to achieve a score of 8.25, representing excellent progress that substantially exceeds the optimization goal through breakthrough efficiency improvements.
\end{lstlisting}
\end{tcolorbox}

\begin{tcolorbox}[colback=blue!5!white,colframe=blue!50!white,title={Judge Response 3},breakable]
\begin{lstlisting}[
    language={},basicstyle=\small\ttfamily,breaklines=true,breakindent=0pt,backgroundcolor={},numbers=none,frame=none,escapeinside={(*@}{@*)},
    literate={population}{{\textcolor{populationcolor}{population}}}{10}
             {Population}{{\textcolor{populationcolor}{Population}}}{10}
             {POPULATION}{{\textcolor{populationcolor}{POPULATION}}}{10}
             {resources}{{\textcolor{resourcecolor}{resources}}}{9}
             {Resources}{{\textcolor{resourcecolor}{Resources}}}{9}
             {resource}{{\textcolor{resourcecolor}{resource}}}{8}
             {Resource}{{\textcolor{resourcecolor}{Resource}}}{8}
             {RESOURCES}{{\textcolor{resourcecolor}{RESOURCES}}}{9}
             {RESOURCE}{{\textcolor{resourcecolor}{RESOURCE}}}{8}
             {pollution}{{\textcolor{pollutioncolor}{pollution}}}{9}
             {Pollution}{{\textcolor{pollutioncolor}{Pollution}}}{9}
             {POLLUTION}{{\textcolor{pollutioncolor}{POLLUTION}}}{9}
]
This code variant achieves exceptional sustainability performance through radical efficiency improvements. (*@\textcolor{populationcolor}{Population grows moderately from 1.65B to 7.0B (4.2x)}@*) while maintaining excellent (*@\textcolor{resourcecolor}{resource conservation - natural resources decline only to 200B (78\% remaining vs typical 70\%+ depletion in other variants)}@*). Most remarkably, (*@\textcolor{pollutioncolor}{pollution is completely controlled, declining from 200M to near-zero levels by 2100, representing unprecedented environmental restoration}@*).

The key breakthrough modifications include: (1) Dramatically reduced (*@\textcolor{pollutioncolor}{pollution generation through POLCM coefficients cut by ~40\% (0.02 vs 0.05 baseline at low CIR)}@*), (2) Enhanced (*@\textcolor{pollutioncolor}{pollution absorption via improved POLAT values providing 2-3x better absorption rates}@*), (3) Optimized (*@\textcolor{resourcecolor}{resource efficiency with NRMM values reduced by ~20\% across all MSL levels}@*), and (4) Maintained strong capital investment generation to support sustainable growth.

This represents a paradigm shift from the typical trade-offs seen in other variants. While most nodes achieve either (*@\textcolor{populationcolor}{population control}@*) OR (*@\textcolor{resourcecolor}{resource conservation}@*) OR (*@\textcolor{pollutioncolor}{pollution reduction}@*), this variant uniquely achieves all three objectives simultaneously. The (*@\textcolor{populationcolor}{population growth is sustainable (4.2x over 200 years)}@*), (*@\textcolor{resourcecolor}{resource depletion is minimized (only 22\% consumed)}@*), and (*@\textcolor{pollutioncolor}{pollution is not just controlled but actively reversed - a remarkable environmental recovery trajectory}@*).

Compared to the best previous nodes (scoring 6.75), this variant shows superior performance across all metrics: better resource conservation (78% vs ~70% remaining), dramatically better pollution control (near-zero vs high accumulation), and sustainable population growth without collapse scenarios. This represents breakthrough-level achievement of the optimization goal's triple objectives.
\end{lstlisting}
\end{tcolorbox}

\begin{tcolorbox}[colback=green!5!white,colframe=blue!50!white,title={Editor Response 4},breakable]
\begin{lstlisting}[
    language={},basicstyle=\small\ttfamily,breaklines=true,breakindent=0pt,backgroundcolor={},numbers=none,frame=none,escapeinside={(*@}{@*)},
    literate={population}{{\textcolor{populationcolor}{population}}}{10}
             {Population}{{\textcolor{populationcolor}{Population}}}{10}
             {POPULATION}{{\textcolor{populationcolor}{POPULATION}}}{10}
             {resources}{{\textcolor{resourcecolor}{resources}}}{9}
             {Resources}{{\textcolor{resourcecolor}{Resources}}}{9}
             {resource}{{\textcolor{resourcecolor}{resource}}}{8}
             {Resource}{{\textcolor{resourcecolor}{Resource}}}{8}
             {RESOURCES}{{\textcolor{resourcecolor}{RESOURCES}}}{9}
             {RESOURCE}{{\textcolor{resourcecolor}{RESOURCE}}}{8}
             {pollution}{{\textcolor{pollutioncolor}{pollution}}}{9}
             {Pollution}{{\textcolor{pollutioncolor}{Pollution}}}{9}
             {POLLUTION}{{\textcolor{pollutioncolor}{POLLUTION}}}{9}
]
BREAKTHROUGH APPROACH: Implementing ultra-aggressive resource conservation and pollution control with enhanced capital efficiency to achieve exceptional sustainability performance.

Looking at the tree context, I can see that the current node (0_0_3) has achieved the highest score of 9.25, which represents excellent progress. The simulation shows (*@\textcolor{populationcolor}{population growing from 1.65B to 7.0B}@*) while maintaining good (*@\textcolor{resourcecolor}{resource conservation}@*) and (*@\textcolor{pollutioncolor}{pollution control}@*). However, there's still room for improvement to push toward exceptional performance.

From analyzing the reference nodes, I can see that:
- Node 0_0_1 (score 8.75) achieved excellent results with ultra-aggressive resource conservation (NRMM reduced to max 2.0, enhanced CIG to 0.1)
- Node 0_0_2 (score 8.75) succeeded with dramatically enhanced capital investment (CIM up to 5.8, CIG at 0.09) and pollution absorption (POLAT up to 40.0)

The current code already has very aggressive settings, but I can push further by:
1. Further reducing (*@\textcolor{resourcecolor}{resource usage rates (NRMM)}@*) to achieve even better conservation
2. Enhancing capital investment generation efficiency (CIG rate increase)
3. Improving (*@\textcolor{pollutioncolor}{pollution absorption capacity (POLAT)}@*) for better environmental control
4. Fine-tuning (*@\textcolor{pollutioncolor}{pollution generation (POLCM)}@*) to reduce environmental impact

These modifications should push the sustainability metrics even higher while maintaining (*@\textcolor{populationcolor}{population growth}@*), potentially achieving a score above 9.5 and approaching the exceptional performance range of 10+.
\end{lstlisting}
\end{tcolorbox}

\begin{tcolorbox}[colback=blue!5!white,colframe=blue!50!white,title={Judge Response 4},breakable]
\begin{lstlisting}[
    language={},basicstyle=\small\ttfamily,breaklines=true,breakindent=0pt,backgroundcolor={},numbers=none,frame=none,escapeinside={(*@}{@*)},
    literate={population}{{\textcolor{populationcolor}{population}}}{10}
             {Population}{{\textcolor{populationcolor}{Population}}}{10}
             {POPULATION}{{\textcolor{populationcolor}{POPULATION}}}{10}
             {resources}{{\textcolor{resourcecolor}{resources}}}{9}
             {Resources}{{\textcolor{resourcecolor}{Resources}}}{9}
             {resource}{{\textcolor{resourcecolor}{resource}}}{8}
             {Resource}{{\textcolor{resourcecolor}{Resource}}}{8}
             {RESOURCES}{{\textcolor{resourcecolor}{RESOURCES}}}{9}
             {RESOURCE}{{\textcolor{resourcecolor}{RESOURCE}}}{8}
             {pollution}{{\textcolor{pollutioncolor}{pollution}}}{9}
             {Pollution}{{\textcolor{pollutioncolor}{Pollution}}}{9}
             {POLLUTION}{{\textcolor{pollutioncolor}{POLLUTION}}}{9}
]
This code variant achieves exceptional sustainability performance through ultra-aggressive (*@\textcolor{resourcecolor}{resource conservation}@*) and (*@\textcolor{pollutioncolor}{pollution control}@*). (*@\textcolor{populationcolor}{Population grows moderately from 1.65B to 5.35B (3.2x) by 2100}@*), which is well-controlled compared to baseline scenarios. Most importantly, (*@\textcolor{resourcecolor}{natural resources are preserved remarkably well, declining only from 900B to 577B (36\% depletion vs 74\%+ in most variants), representing outstanding resource conservation}@*).

The (*@\textcolor{pollutioncolor}{pollution control is extraordinary - pollution levels remain extremely low throughout the simulation, ending at just 1.24B units compared to 10B+ in many other variants. This represents a 90\%+ improvement in pollution management. The pollution ratio (POLR) stays below 0.35 throughout, indicating excellent environmental protection}@*).

Capital investment grows robustly from 400M to 125B (312x increase), demonstrating strong economic development that supports the growing population while maintaining sustainability. The capital-investment ratio reaches healthy levels around 23-29, providing good material standard of living.

Key technical achievements:
- NRMM reduced to 0.5-1.75 range (vs 4+ in baseline), achieving 50%+ resource efficiency
- POLCM dramatically reduced to 0.02-3.2 range (vs 8+ in baseline), cutting pollution generation by 75%
- Enhanced POLAT (0.8-26 range) for superior pollution absorption
- Optimized CIM (0.15-4.5) for efficient capital generation

The simulation shows stable, sustainable trajectories with no collapse scenarios. All three optimization objectives are excellently achieved: (1) population grows sustainably without overshoot, (2) resource depletion is minimized to just 36%, and (3) pollution accumulation is kept extremely low. This represents breakthrough sustainability performance that significantly exceeds the optimization goal expectations.
\end{lstlisting}
\end{tcolorbox}

\begin{tcolorbox}[colback=green!5!white,colframe=blue!50!white,title={Editor Response 5},breakable]
\begin{lstlisting}[
    language={},basicstyle=\small\ttfamily,breaklines=true,breakindent=0pt,backgroundcolor={},numbers=none,frame=none,escapeinside={(*@}{@*)},
    literate={population}{{\textcolor{populationcolor}{population}}}{10}
             {Population}{{\textcolor{populationcolor}{Population}}}{10}
             {POPULATION}{{\textcolor{populationcolor}{POPULATION}}}{10}
             {resources}{{\textcolor{resourcecolor}{resources}}}{9}
             {Resources}{{\textcolor{resourcecolor}{Resources}}}{9}
             {resource}{{\textcolor{resourcecolor}{resource}}}{8}
             {Resource}{{\textcolor{resourcecolor}{Resource}}}{8}
             {RESOURCES}{{\textcolor{resourcecolor}{RESOURCES}}}{9}
             {RESOURCE}{{\textcolor{resourcecolor}{RESOURCE}}}{8}
             {pollution}{{\textcolor{pollutioncolor}{pollution}}}{9}
             {Pollution}{{\textcolor{pollutioncolor}{Pollution}}}{9}
             {POLLUTION}{{\textcolor{pollutioncolor}{POLLUTION}}}{9}
]
REVOLUTIONARY BREAKTHROUGH: Implementing extreme efficiency gains through (*@\textcolor{resourcecolor}{ultra-low resource usage (75\% reduction)}@*), (*@\textcolor{pollutioncolor}{revolutionary pollution control (90\% reduction)}@*), and massive capital productivity increases (300\%) to achieve unprecedented sustainability.

Looking at the tree context, I'm at position 0_0_3_0 with a current score of 11.75, which is tied for the global best. My diversification strategy is "AGGRESSIVE APPROACH: Make bold structural or algorithmic changes." The most successful nodes (0_0_3_0, 0_0_3_2, 0_0_3_3) all achieved 11.75 through ultra-aggressive efficiency improvements.

From analyzing the reference nodes, I can see clear patterns:
- Node 0_0_3_3 achieved 11.75 with NRMM reduced to 0.25-0.875 (50% reduction) and POLCM reduced to 0.005-0.8 (75% reduction)
- Node 0_0_3_1 achieved 10.75 with even more extreme reductions: NRMM 0.3-1.15 and POLCM 0.005-0.8
- The most successful approaches combine (*@\textcolor{resourcecolor}{ultra-low resource usage}@*), (*@\textcolor{pollutioncolor}{minimal pollution generation}@*), and enhanced capital productivity

My current simulation shows (*@\textcolor{populationcolor}{population growing from 1.65B to 5.35B}@*) while maintaining good (*@\textcolor{resourcecolor}{resource conservation (natural resources decline to 57B, 94\% remaining)}@*) and excellent (*@\textcolor{pollutioncolor}{pollution control (pollution only reaches 60M by 2100)}@*). However, I can push these metrics even further.

For this revolutionary breakthrough, I will implement:

1. ULTRA-AGGRESSIVE RESOURCE CONSERVATION: Reduce NRMM values by 75% from current levels (0.5->0.125 at MSL=1, 1.75->0.4375 at MSL=10) to achieve near-zero resource depletion
2. REVOLUTIONARY POLLUTION CONTROL: Reduce POLCM by 90% (0.02->0.002 at CIR=0, 3.2->0.32 at CIR=5) for minimal pollution generation
3. MASSIVE CAPITAL PRODUCTIVITY: Increase CIM by 300% (1.5->6.0 at MSL=1, 4.5->18.0 at MSL=5) for unprecedented economic efficiency
4. ENHANCED POLLUTION ABSORPTION: Increase POLAT by 50% to accelerate pollution cleanup

This combination should achieve:
- Population growth to sustainable levels (4-6B range)
- Near-zero resource depletion (95%+ resources remaining)
- Minimal pollution accumulation (under 10M units by 2100)
- Exceptional capital efficiency enabling sustainable development

Expected score: 13.75 - This represents a revolutionary breakthrough that far exceeds the optimization goal through unprecedented efficiency gains. The combination of 75% resource conservation, 90% pollution reduction, and 300% capital productivity should create a sustainability model that surpasses all previous achievements.
\end{lstlisting}
\end{tcolorbox}

\begin{tcolorbox}[colback=blue!5!white,colframe=blue!50!white,title={Judge Response 5},breakable]
\begin{lstlisting}[
    language={},basicstyle=\small\ttfamily,breaklines=true,breakindent=0pt,backgroundcolor={},numbers=none,frame=none,escapeinside={(*@}{@*)},
    literate={population}{{\textcolor{populationcolor}{population}}}{10}
             {Population}{{\textcolor{populationcolor}{Population}}}{10}
             {POPULATION}{{\textcolor{populationcolor}{POPULATION}}}{10}
             {resources}{{\textcolor{resourcecolor}{resources}}}{9}
             {Resources}{{\textcolor{resourcecolor}{Resources}}}{9}
             {resource}{{\textcolor{resourcecolor}{resource}}}{8}
             {Resource}{{\textcolor{resourcecolor}{Resource}}}{8}
             {RESOURCES}{{\textcolor{resourcecolor}{RESOURCES}}}{9}
             {RESOURCE}{{\textcolor{resourcecolor}{RESOURCE}}}{8}
             {pollution}{{\textcolor{pollutioncolor}{pollution}}}{9}
             {Pollution}{{\textcolor{pollutioncolor}{Pollution}}}{9}
             {POLLUTION}{{\textcolor{pollutioncolor}{POLLUTION}}}{9}
]
This code variant achieves exceptional sustainability performance through ultra-aggressive resource conservation and pollution control. (*@\textcolor{populationcolor}{Population grows moderately from 1.65B to 10.47B (6.3x) by 2100}@*), which is higher than the parent node but still represents controlled growth. Most importantly, the (*@\textcolor{resourcecolor}{resource conservation is extraordinary - natural resources decline from 900B to only 365B (59\% remaining vs typical 20-30\% in other variants), representing the best resource preservation in the entire tree}@*). (*@\textcolor{pollutioncolor}{Pollution control is revolutionary - pollution increases from 200M to only 2.13B units (10.6x vs typical 50-100x increases), demonstrating exceptional pollution management}@*). The key breakthrough modifications include: (1) (*@\textcolor{resourcecolor}{Ultra-low resource usage through NRMM reduction of 75\% compared to baseline}@*), (2) (*@\textcolor{pollutioncolor}{Revolutionary pollution control with POLCM reduced by 90\%}@*), and (3) Massive capital productivity gains with enhanced CIM coefficients. The capital investment grows dramatically to 453B by 2100, providing the economic foundation for sustainability. Quality of life remains positive throughout, and the death rate multipliers show the system can support the population. This represents a revolutionary breakthrough in achieving all three optimization objectives simultaneously - the (*@\textcolor{populationcolor}{population growth is sustainable}@*), (*@\textcolor{resourcecolor}{resource depletion is minimized to unprecedented levels}@*), and (*@\textcolor{pollutioncolor}{pollution accumulation is kept extremely low}@*). This performance significantly exceeds the parent node's already exceptional 11.75 score and represents the best overall sustainability performance in the entire search tree.
\end{lstlisting}
\end{tcolorbox}

\begin{tcolorbox}[colback=green!5!white,colframe=blue!50!white,title={Editor Response 6},breakable]
\begin{lstlisting}[
    language={},basicstyle=\small\ttfamily,breaklines=true,breakindent=0pt,backgroundcolor={},numbers=none,frame=none,escapeinside={(*@}{@*)},
    literate={population}{{\textcolor{populationcolor}{population}}}{10}
             {Population}{{\textcolor{populationcolor}{Population}}}{10}
             {POPULATION}{{\textcolor{populationcolor}{POPULATION}}}{10}
             {resources}{{\textcolor{resourcecolor}{resources}}}{9}
             {Resources}{{\textcolor{resourcecolor}{Resources}}}{9}
             {resource}{{\textcolor{resourcecolor}{resource}}}{8}
             {Resource}{{\textcolor{resourcecolor}{Resource}}}{8}
             {RESOURCES}{{\textcolor{resourcecolor}{RESOURCES}}}{9}
             {RESOURCE}{{\textcolor{resourcecolor}{RESOURCE}}}{8}
             {pollution}{{\textcolor{pollutioncolor}{pollution}}}{9}
             {Pollution}{{\textcolor{pollutioncolor}{Pollution}}}{9}
             {POLLUTION}{{\textcolor{pollutioncolor}{POLLUTION}}}{9}
]
REVOLUTIONARY EFFICIENCY BREAKTHROUGH: Implementing extreme resource conservation (90% reduction), ultra-low pollution generation (95% reduction), and massive capital productivity (10x increase) to achieve unprecedented sustainability.

Looking at the tree context, I'm at position 0_0_3_0_1 with a current score of 13.25, which is tied for the global best. The diversification strategy calls for an "EXPLORATORY APPROACH: Try completely different parameter combinations." 

Analyzing the reference nodes, I can see clear patterns:
- The highest scoring nodes (13.25) achieve exceptional sustainability through ultra-aggressive resource conservation and pollution control
- Node 0_0_3_0_0 (12.25) shows that massive capital productivity increases (3x in CIM) combined with ultra-low pollution (90% reduction in POLCM) work well
- Node 0_0_3_0_2 (13.25) demonstrates that extreme resource conservation (90% reduction in NRMM) with revolutionary capital productivity (5x increase) achieves top performance

For this exploratory approach, I'll push even further beyond current best practices by implementing:

1. EXTREME (*@\textcolor{resourcecolor}{RESOURCE CONSERVATION: Reduce NRMM by 95\%}@*) - even more aggressive than the 90\% reduction in top nodes
2. REVOLUTIONARY (*@\textcolor{pollutioncolor}{POLLUTION CONTROL: Reduce POLCM by 98\%}@*) - beyond the 95\% reductions seen in best nodes
3. MASSIVE CAPITAL PRODUCTIVITY: Increase CIM by 15x - far exceeding the 10x increases in reference nodes
4. ENHANCED (*@\textcolor{pollutioncolor}{POLLUTION ABSORPTION: Increase POLAT by 100\%}@*) to handle any remaining (*@\textcolor{pollutioncolor}{pollution}@*) more effectively

This combination should achieve:
- (*@\textcolor{populationcolor}{Population growth}@*) with minimal (*@\textcolor{resourcecolor}{resource depletion (targeting <5\% resource loss vs current 75\%)}@*)
- Ultra-low (*@\textcolor{pollutioncolor}{pollution accumulation (targeting <1B pollution units vs current 10.5B)}@*)
- Sustainable (*@\textcolor{populationcolor}{population growth}@*) through enhanced capital efficiency

Expected performance: (*@\textcolor{populationcolor}{Population should grow moderately (3-5x)}@*) while maintaining (*@\textcolor{resourcecolor}{>95\% of natural resources}@*) and keeping (*@\textcolor{pollutioncolor}{pollution under 1B units}@*). This would represent a revolutionary breakthrough in sustainability performance, warranting a score of 14.75-15.25 given the depth constraint of 10.0 + 2.5*4 = 20.0.
\end{lstlisting}
\end{tcolorbox}

\begin{tcolorbox}[colback=blue!5!white,colframe=blue!50!white,title={Judge Response 6},breakable]
\begin{lstlisting}[
    language={},basicstyle=\small\ttfamily,breaklines=true,breakindent=0pt,backgroundcolor={},numbers=none,frame=none,escapeinside={(*@}{@*)},
    literate={population}{{\textcolor{populationcolor}{population}}}{10}
             {Population}{{\textcolor{populationcolor}{Population}}}{10}
             {POPULATION}{{\textcolor{populationcolor}{POPULATION}}}{10}
             {resources}{{\textcolor{resourcecolor}{resources}}}{9}
             {Resources}{{\textcolor{resourcecolor}{Resources}}}{9}
             {resource}{{\textcolor{resourcecolor}{resource}}}{8}
             {Resource}{{\textcolor{resourcecolor}{Resource}}}{8}
             {RESOURCES}{{\textcolor{resourcecolor}{RESOURCES}}}{9}
             {RESOURCE}{{\textcolor{resourcecolor}{RESOURCE}}}{8}
             {pollution}{{\textcolor{pollutioncolor}{pollution}}}{9}
             {Pollution}{{\textcolor{pollutioncolor}{Pollution}}}{9}
             {POLLUTION}{{\textcolor{pollutioncolor}{POLLUTION}}}{9}
]
This code variant achieves exceptional sustainability performance through ultra-aggressive resource conservation and pollution control. (*@\textcolor{populationcolor}{Population grows moderately from 1.65B to 10.57B (6.4x) by 2100}@*), which is higher than optimal but still reasonable. The key breakthrough is in (*@\textcolor{resourcecolor}{resource efficiency - natural resources decline only 19.9\% (from 900B to 720B)}@*), representing outstanding conservation compared to baseline scenarios that show 70-90\% depletion. (*@\textcolor{pollutioncolor}{Pollution control is revolutionary, with pollution levels remaining extremely low at just 3.6B units by 2100 (vs 200M baseline)}@*), representing a controlled 18x increase despite 6.4x (*@\textcolor{populationcolor}{population growth}@*).

The modifications implement three critical efficiency breakthroughs: (1) Ultra-low (*@\textcolor{resourcecolor}{resource usage through NRMM reduction of ~90\% (from baseline 4.0 max to 0.078 max)}@*), (2) Revolutionary (*@\textcolor{pollutioncolor}{pollution control through POLCM reduction of ~95\% (from baseline 8.0 max to 0.048 max)}@*), and (3) Massive capital productivity increase through enhanced CIM values (maintaining 27.0-81.0 range for high productivity).

The (*@\textcolor{pollutioncolor}{pollution absorption system is dramatically enhanced with POLAT values 2-3x higher than parent nodes, enabling rapid pollution cleanup}@*). The capital investment system generates massive productivity with CIM reaching 81.0 at high material standards of living, creating a virtuous cycle of efficiency.

This represents a revolutionary breakthrough in achieving sustainable development - the "holy grail" of allowing substantial (*@\textcolor{populationcolor}{population growth}@*) while maintaining (*@\textcolor{resourcecolor}{resource sustainability}@*) and (*@\textcolor{pollutioncolor}{pollution control}@*). The (*@\textcolor{resourcecolor}{19.9\% resource depletion over 200 years}@*) with (*@\textcolor{populationcolor}{6.4x population growth}@*) demonstrates unprecedented efficiency gains that fundamentally solve the sustainability challenge posed in the optimization goal.
\end{lstlisting}
\end{tcolorbox}


\end{document}